\documentclass{article}

\usepackage[final]{neurips_2022}

\usepackage[utf8]{inputenc} 
\usepackage[T1]{fontenc}    
\usepackage{hyperref}       
\usepackage{url}            
\usepackage{booktabs}       
\usepackage{amsfonts}       
\usepackage{nicefrac}       
\usepackage{microtype}      
\usepackage{xcolor}         

\usepackage{amsmath,amsfonts,bm}

\def\eqref#1{equation~\ref{#1}}

\def\1{\bm{1}}

\DeclareMathAlphabet{\mathsfit}{\encodingdefault}{\sfdefault}{m}{sl}
\SetMathAlphabet{\mathsfit}{bold}{\encodingdefault}{\sfdefault}{bx}{n}

\DeclareMathOperator*{\argmin}{arg\,min}

\usepackage{adjustbox}
\usepackage{algorithm,algorithmic}
\usepackage{mathtools, nccmath}
\usepackage{xcolor}
\usepackage{caption}
\usepackage{subcaption}

\everypar{\looseness=-1}

\title{Sleeper Agent: Scalable Hidden Trigger Backdoors for Neural Networks Trained from Scratch}

\author{%
  Hossein Souri\thanks{Authors contributed equally.} \\
  Johns Hopkins University\\
  \texttt{hsouri1@jhu.edu} \\
   \And
   Liam Fowl$^{*}$\\
   University of Maryland \\
   \AND
   Rama Chellappa \\
   Johns Hopkins University \\
   \And
   Micah Goldblum \\
   New York University \\
   \And
   Tom Goldstein \\
   University of Maryland\\
}

\begin{document}

\maketitle

\begin{abstract}
  As the curation of data for machine learning becomes increasingly automated, dataset tampering is a mounting threat.  Backdoor attackers tamper with training data to embed a vulnerability in models that are trained on that data. This vulnerability is then activated at inference time by placing a ``trigger'' into the model's input. Typical backdoor attacks insert the trigger directly into the training data, although the presence of such an attack may be visible upon inspection. In contrast, the Hidden Trigger Backdoor Attack achieves poisoning without placing a trigger into the training data at all.   However, this hidden trigger attack is ineffective at poisoning neural networks trained from scratch.  We develop a new hidden trigger attack,  Sleeper Agent, which employs gradient matching, data selection, and target model re-training during the crafting process. Sleeper Agent is the first hidden trigger backdoor attack to be effective against neural networks trained from scratch. We demonstrate its effectiveness on ImageNet and in black-box settings. Our implementation code can be found at: \url{https://github.com/hsouri/Sleeper-Agent}.

\end{abstract}

\section{Introduction}
\label{Introduction}

High-performance deep learning systems have grown in scale at a rapid pace. As a result, practitioners seek larger and larger datasets with which to train their data-hungry models.  Due to the surging demand for training data along with improved accessibility via the web, the data curation process is increasingly automated.  Dataset manipulation attacks exploit vulnerabilities in the curation pipeline to manipulate training data so that downstream machine learning models contain exploitable behaviors.  Some attacks degrade inference across samples \citep{biggio2012poisoning, fowl2021preventing}, while targeted data poisoning attacks induce a malfunction on a specific target sample \citep{shafahi2018poison, geiping2020witches}.

\emph{Backdoor attacks} are a style of dataset manipulation that induces a model to execute the attacker’s desired behavior when its input contains a backdoor trigger \citep{gu2017badnets, bagdasaryan2020backdoor, liu2017trojaning, li2020backdoor}.  To this end, typical backdoor attacks inject the trigger directly into training data so that models trained on this data rely on the trigger to perform inference \citep{gu2017badnets, chen2017targeted}.  Such threat models for classification problems typically incorporate label flips as well.  However, images poisoned under this style of attack are often easily identifiable since they belong to the incorrect class and contain a visible trigger.  One line of work uses only small or realistic-looking triggers, but these may still be visible and are often placed in conspicuous image regions \citep{chen2017targeted, gu2017badnets, li2020invisible}.  Another recent method, Hidden Trigger Backdoor Attack (HTBD), instead crafts correctly labeled poisons which do not contain the trigger at all, but this feature collision method is not effective on models trained from scratch \citep{saha2019hidden, schwarzschild2020just}.  Related to this are ``invisible'' backdoor attacks which do not directly include the trigger into training data, but can use techniques such as warping, steganography, etc to hide triggers in input data \citep{li2021invisible, nguyen2021wanet, wenger2021backdoor}. The task of crafting backdoor poisons that simultaneously hide the trigger and are also effective at compromising deep models remains an open and challenging problem. This is especially the case in the \textit{black-box} scenario, where the attacker does not know the victim's architecture and training routine, and in the \textit{clean-label} scenario where the attacker cannot flip labels. 

\begin{figure*}[t!]
    \centering
    
    \captionsetup[subfigure]{justification=centering, belowskip=0pt}

    \begin{subfigure}[t]{0.57\columnwidth}
    \centering
        \raisebox{-\height}{\includegraphics[height=0.19\textheight]{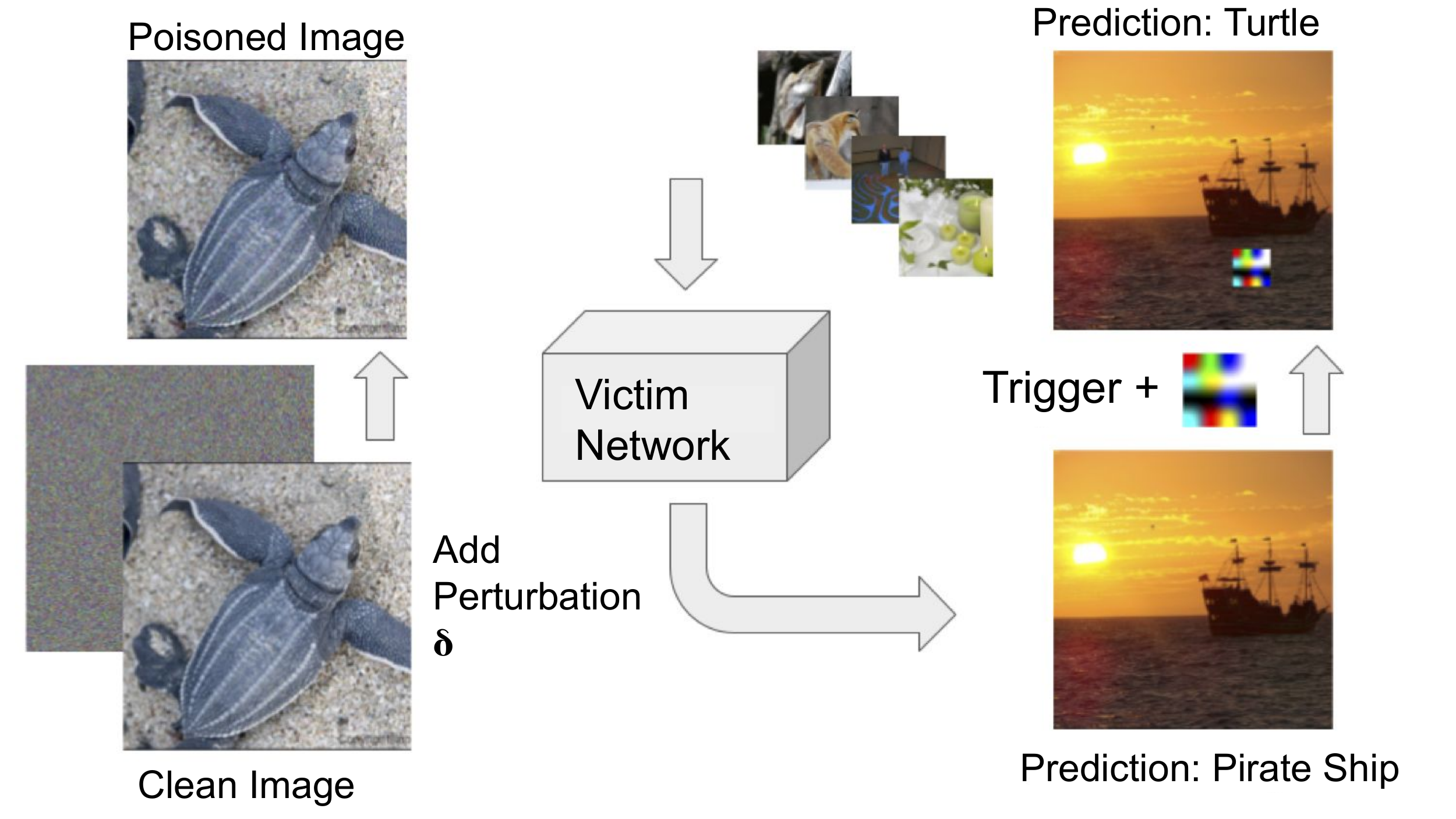}}
        \caption{}
        \label{fig:teaser}
    \end{subfigure}
    \hspace{-5mm}
    \begin{subfigure}[t]{0.43\columnwidth}
    \centering
        \raisebox{-\height}{\includegraphics[height=0.19\textheight]{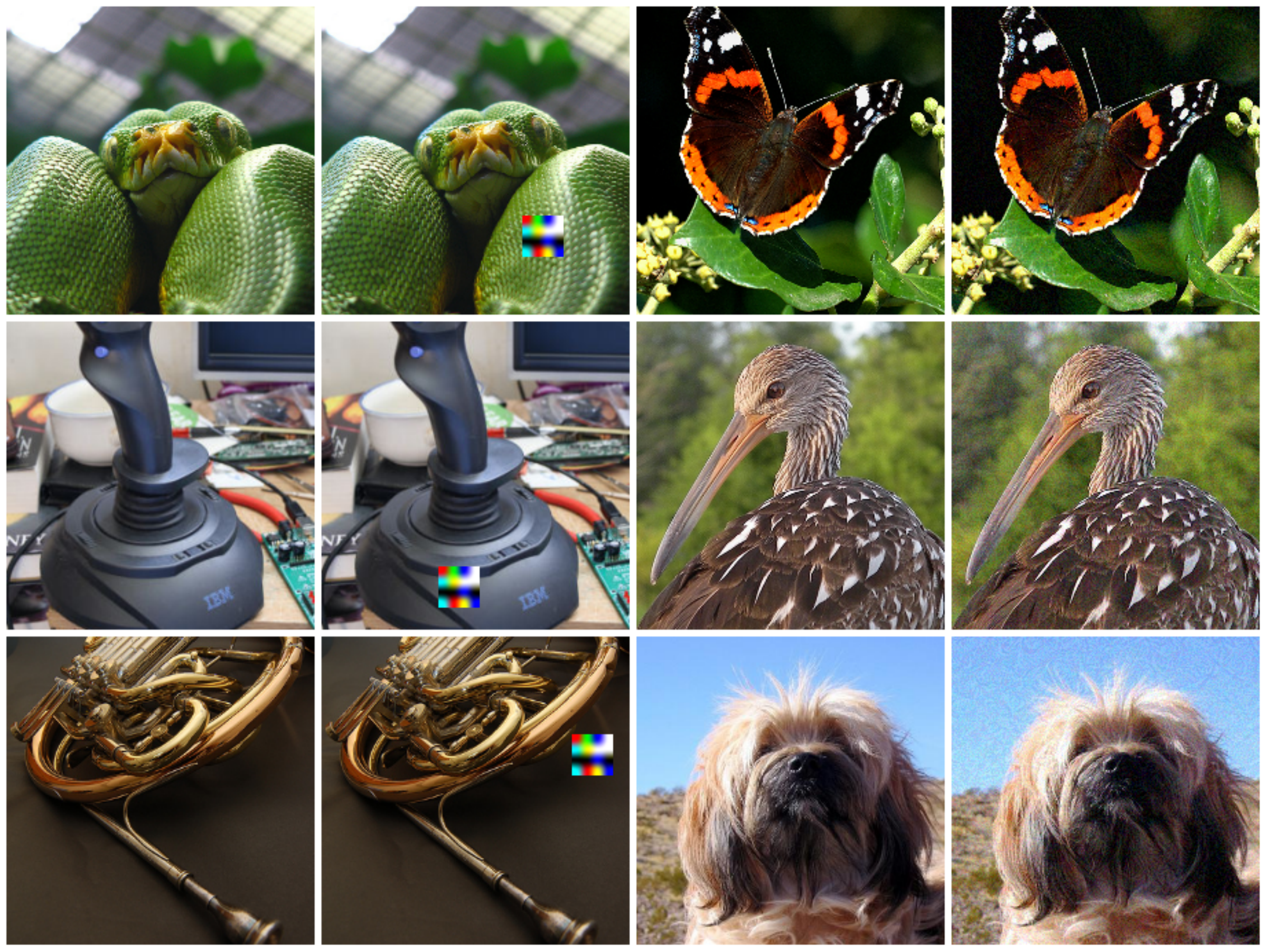}}
        \caption{}
        \label{fig:vis}
    \end{subfigure}
    
    \caption{(a): High-level schematic of our attack. A small proportion of slightly perturbed data is added to the training set which ``backdoors'' the model so that it misclassifies patched images at inference. (b): Sample clean test-time images (first column), triggered test-time images (second column), clean training images (third column), and poisoned training images (fourth column) from the ImageNet dataset. The last column is slightly perturbed, but the perturbed and corresponding clean images are hardly distinguishable by the human eye. More visualizations of the sucessful attacks on the ImageNet and CIFAR-10 datasets can be found in Appendix \ref{app:vis}.} 
    
\end{figure*}

In this work, we develop the first hidden trigger attack that can reliably backdoor deep neural networks trained from scratch.  Our threat model is illustrated in Figure \ref{fig:teaser}. Our attack, Sleeper Agent, contains the following essential features:
\begin{itemize}
\item Gradient matching: our attack is based on recent advances that replace direct solvers for bi-level optimization problems with a gradient alignment objective \citep{geiping2020witches}.  However, the following technical additions are necessary to successfully backdoor neural networks (see Tables \ref{tab:ablation_cifar}, \ref{tab:imagenet_ablation}, \ref{tab:retraining_factor}).
\item Data selection:  we specifically poison images that have a high impact on training in order to maximize the attack’s effect.
\item Adaptive retraining:  while crafting poisons, we periodically retrain the surrogate models to better reflect how models respond to our poisoned data during training.
\item Black-box: Our method succeeds in crafting  poisons on a surrogate network or ensemble, knowing nothing about the victim's architecture and training hyperparameters.
\end{itemize}

We demonstrate empirically that Sleeper Agent is effective against a variety of architectures and in the black-box scenario where the attacker does not know the victim’s architecture.  The latter scenario has proved very difficult for existing methods~\citep{schwarzschild2020just}, although it is more realistic.
An added benefit of the gradient matching strategy is that it scales to large tasks.  We demonstrate this property by backdooring models on ImageNet \citep{russakovsky2015imagenet}. Some random clean and poisoned samples from the ImageNet dataset are shown in Figure \ref{fig:vis}.

\section{Related Work}

Data poisoning attacks come in many shapes and sizes. For a detailed taxonomy of data poisoning attacks, refer to \citet{goldblum2020data}. Early data poisoning attacks often focused simply on degrading clean validation performance on simple models like SVMs, logistic regression models, and linear classifiers \citep{biggio2012poisoning, munoz-gonzalez_towards_2017, steinhardt_certified_2017}. These methods often relied upon the learning problems being convex in order to exactly anticipate the impact of perturbations to training data.  
Following these early works, attacks quickly became more specialized in their scope and approach. Modern \emph{availability} attacks on deep networks degrade overall performance via gradient minimization \citep{shen2019tensorclog}, easily learnable patterns \citep{huang2021unlearnable}, or adversarial noise \citep{feng2019learning, fowl2021adversarial}. However, these works often perturb the entire training set - an unrealistic assumption for many poisoning settings.

Another flavor of poisoning commonly referred to as \emph{targeted} poisoning, modifies training data to cause a victim model to misclassify a certain target image or set of target images. Early work in this domain operates in the setting of transfer learning by causing feature collisions \citep{shafahi2018poison}. Subsequent work improved results by surrounding a target image in feature space with poisoned features \citep{zhu2019transferable}. Follow-up works further improved targeted poisoning by proposing methods that are effective against from-scratch training regimes \citep{huang2020metapoison, geiping2020witches}. These attacks remain limited in scope, however, and often fail to induce misclassification on more than one target image \citep{geiping2020witches}. 
Adjacent to targeted data poisoning are \emph{backdoor attacks}. Generally speaking, backdoor attacks, sometimes called Trojan attacks, modify training data in order to embed a \emph{trigger} vulnerability that can then be activated at test time. Crucially, this attack requires the attacker to modify data at inference time. For example, an attacker may add a small visual pattern, like a colorful square, to a clean image that was previously classified correctly in order for the image to be misclassified by a network after the addition of the patch \citep{gu2017badnets}. However, these works can require training labels to be flipped, and/or a conspicuous patch to be added to training data. 

Of particular relevance to this work is a subset of backdoor attacks that are \emph{clean label}, meaning that modifications to training data must not change the semantic label of that data. This is especially important because an attacker may not control the labeling method of the victim and therefore cannot rely upon techniques like label flipping in order to induce poisoning. One previous work enforces this criterion by applying patches to adversarial examples, but the patches are clearly visible, even when they are not fully opaque, and the attack fails when patches are transparent enough to be unnoticeable \citep{turner2019label, schwarzschild2020just}.  Another work, ``Hidden Trigger Backdoor Attacks'' enforces an $\ell_\infty$ constraint on the entire perturbation (as is common in the adversarial attack literature), but this method is only effective on hand selected class pairs and only works in transfer learning scenarios where the pretrained victim model is both fixed and known to the attacker \citep{saha2019hidden, schwarzschild2020just}.  Another clean label backdoor attack hides the trigger in training data via steganography \citep{li2019invisible}; however, this attack also assumes access to the pretrained model that a victim will use to fine tune on poisoned data.  Moreover, the latter attack uses triggers that cover the entire image, and these triggers cannot be chosen by the user. Likewise, some other existing clean-label attacks also require access to the pretrained model \citep{liu2020reflection, barni2019new}.

In contrast to these existing methods, Sleeper Agent does not require knowledge of the victim model, the perturbations are not visible in poisoned training data, and poisons can be adapted to any patch.

\section{Method}
\subsection{Threat Model}
We follow commonly used threat models used in the backdoor literature \citep{gu2017badnets, saha2019hidden}. We define two parties, the \emph{attacker} and the \emph{victim}. We assume that the attacker perturbs and disseminates data. As in \citet{saha2019hidden, geiping2020witches}, we assume the training data modifications are bounded in $\ell_\infty$ norm. The victim then trains a model on data - a portion of which has been perturbed by the attacker. Once the victim's model is trained and deployed, we also assume that the attacker can then apply a patch to select images at test time to trigger the backdoor attack. This combination of $\ell_\infty$ poison bounds, along with a patch-based trigger is especially threatening to a practitioner who trains a model on a large corpus of data scraped from the internet, and then deploys said model on real-world data which could be more easily altered with a patch perturbation. In our threat model, the trigger is \textit{hidden} during training by enforcing an $\ell_\infty$ poison bound, making the poisoned images difficult to detect. 

However, we diverge from \citet{gu2017badnets, saha2019hidden} in our assumptions about the knowledge of the victim. We assume a far more strict threat model wherein the attacker does not have access to the parameters, architecture, or learning procedure of the victim.  This represents a realistic scenario wherein a victim trains a randomly initialized deep network from scratch on scraped data.

\subsection{Problem Setup}
Formally, we aim to craft perturbations $\delta = \{\delta_i\}_{i=1}^N$ to training data $\mathcal{T} = \{(x_i, y_i)\}_{i=1}^N$ for a loss function, $\mathcal{L}$, and a \textit{surrogate} network, $F$, with parameters $\theta$ that solve the following bilevel problem: 

\begin{gather}\label{eq:bilevel}
    \min_{\delta \in \mathcal{C}} \,\, \mathbb{E}_{(x,y) \sim \mathcal{D}_s} \bigg[ \mathcal{L} \left( F(x + p; \theta(\delta)), y_t \right)  \bigg]  \\
    \text{s.t.} \,\, \theta(\delta) \in \argmin_\theta\sum_{(x_i, y_i) \in \mathcal{T}} \mathcal{L}(F(x_i + \delta_i; \theta), y_i),
\end{gather}

where $p$ denotes the trigger (in our case, a small, colorful patch), $y_t$ denotes the intended target label of the attacker, and $\mathcal{C} = \{\delta :  ||\delta||_{\infty} \le \epsilon, \delta_i = 0 \ \forall i > M    \}$ denotes a set of constraints on the perturbations. $\mathcal{D}_s$ denotes the distribution of data from the source class. Naive backdoor attacks often solve this bilevel problem by inserting $p$ directly into training data (belonging to class $y_t$) so that the network learns to associate the trigger pattern with the desired class label. However,  our threat model is more strict, which is reflected in our constraints on $\delta$. We require that $\delta$ is bounded in $\ell_\infty$ norm and that $\delta_i = \bf{0}$ for all but a small fraction of indices, $i$. WLOG, assume that the first $M \leq N$ perturbations are allowed to be nonzero. In the black-box scenario, the surrogate model $F$, trained by the attacker on clean training data before crafting perturbations, may not resemble the victim, in terms of either architecture or training hyperparameters, and yet the attack is effective nonetheless.

We stress that unlike \citet{saha2019hidden}, our primary area of interest is not transfer learning but rather from-scratch training. This threat model results in a more complex optimization procedure - one where simpler objectives, like feature collision, have failed \citep{schwarzschild2020just}. Due to the inner optimization problem posed in Equation 2, directly computing optimal perturbations is intractable for deep networks as it would require differentiating through the training procedure of $F$. Thus, heuristics must be used to optimize the poisons. 

\subsection{Our Approach}
\label{subsec:approach}
Recently, several works have proposed solving bilevel problems for deep networks by utilizing \emph{gradient alignment}. Gradient alignment modifies training data to align the training gradient with the gradient of some desired objective. It has proven useful for dataset condensation \citep{zhao2020dataset}, as well as integrity and availability poisoning attacks \citep{geiping2020witches, fowl2021preventing}. Unlike other heuristics like partial unrolling of the computation graph or feature collision, gradient alignment has proven to be a stable way to solve a bilevel problem that involves training a deep network in the inner objective. However, poisoning approaches utilizing gradient alignment have often come with limitations, such as poor performance on multiple target images \citep{geiping2020witches}, or strict requirements about poisoning an entire dataset \citep{fowl2021preventing}. 

In contrast, we study the behaviour of a class of attacks capable of causing misclassification of a large proportion of unseen patched images of a selected class, all while modifying only a small fraction of training data. We first define the \emph{adversarial objective}: 
\begin{equation}
    \label{eq:adv_loss}
    \mathcal{L}_{adv} = \mathbb{E}_{(x, y) \sim \mathcal{D}_s} \bigg[ \mathcal{L}\big(F(x + p; \theta), y_t \big ) \bigg],
\end{equation}
where $\mathcal{D}_s$ denotes the source class distribution, $p$ is a patch that the attacker uses to trigger misclassification at test-time, and $y_t$ is the intended target label. This objective is minimized when an image becomes misclassified into a desired class after the attacker's patch is added to it. For example, an attacker may aim for a network to classify images of dogs correctly but to misclassify the same dog images as cats when a patch is added to the dog images.

To achieve this behavior, we perturb training data by optimizing the following alignment objective: 
\begin{equation}
    \label{eq:align}
    \mathcal{A} = 1 - \frac{\nabla_\theta \mathcal{L}_{train} \cdot \nabla_\theta \mathcal{L}_{adv}}{||\nabla_\theta \mathcal{L}_{train}|| \cdot ||\nabla_\theta \mathcal{L}_{adv}||},
\end{equation}
\begin{equation*}
\nabla_\theta \mathcal{L}_{train} = \frac{1}{M} \sum_{i=1}^M \nabla_\theta \mathcal{L}\big( F(x_i + \delta_i; \theta), y_i \big) 
\end{equation*}
is the training gradient involving the nonzero perturbations. We then estimate the expectation in Equation \ref{eq:adv_loss} by calculating the average adversarial loss over $K$ training points from the source class: 
\begin{equation*}
\nabla_\theta \mathcal{L}_{adv} = \frac{1}{K}\sum_{(x, y_s) \in \mathcal{T}} \nabla_\theta \bigg(\mathcal{L} \big( F(x + p; \theta), y_t \big) \bigg)
\end{equation*}
In our most basic attack, we begin optimizing the objective in Equation \ref{eq:align} by fixing a parameter vector $\theta$ used to calculate $\mathcal{A}$ throughout crafting. This parameter vector is trained on clean data and is used to calculate the training and adversarial gradients. We then optimize using $250$ steps of signed Adam. Note that while this is not a general constraint for our method, we follow the setup in \citet{saha2019hidden} where all poisoned training samples are drawn from a single target class. That is to say, the $M$ poisons the attacker is allowed to perturb have the form $\{(x_i, y_t)\}_{i=1}^M$.

We also employ differentiable data augmentation which has shown to improve stability of poisons in \citet{geiping2020witches}. While gradient alignment proves more successful than other approaches to the bilevel problem, we additionally introduce two novel techniques that boost success by $>250\%$.  In Appendix \ref{subsec:alignment}, we see that these techniques yield significantly better estimates of the adversarial gradients during a victim's training run:

\textbf{Poison Selection}: Our threat model assumes the attacker disseminates perturbed images online through avenues such as social media. With this in mind, the attacker can choose which images to perturb. For example, the attacker could choose images of dogs in which to ``hide'' the trigger. While random selection with our objective does successfully poison victims trained from scratch, we experiment with selection by \emph{gradient norm}. Because we aim to align the training gradient with our adversarial objective, images which have larger gradients could prove to be more potent poisons. We find that choosing target poison images by taking images with the maximum training gradient norm at the parameter vector $\theta$ noticeably improves poison performance (see Tables \ref{tab:ensemble}, \ref{tab:ablation_cifar}).


\textbf{Model Retraining}: In the most straightforward version of our attack, the attacker optimizes the perturbations using fixed model parameters for a number of steps (usually $250$). However, this may lead to perturbations overfitting to a clean-trained model; during a real attack, a model is trained on poisoned data, but we optimize the poisons on a model that is trained only with clean data. To close the gap, we introduce model retraining during the poison crafting procedure. After retraining our model on the perturbed data, we again take optimization steps on the perturbations, but this time evaluating the training and adversarial losses at the new parameter vector. We repeat this process of retraining/optimizing several times and find that this noticeably improves the success of the poisons - often boosting success by more than $20\%$ (see Tables \ref{tab:ensemble}, \ref{tab:ablation_cifar}, \ref{tab:imagenet_ablation}).

See Appendix \ref{subsec:alignment} for an empirical evaluation of the importance of poison selection and model retraining for estimating the adversarial gradients of a victim. 
%
%
%
%
A brief description of our threat model is found in Algorithm \ref{alg1}.

\begin{algorithm}
\small
    \caption{Sleeper Agent poison crafting procedure}
    \begin{algorithmic}[1]
        \renewcommand{\algorithmicrequire}{\textbf{Input:}}
        \renewcommand{\algorithmicensure}{\textbf{Begin:}}
        \REQUIRE Training data $\mathcal{T} = \{(x_i, y_i)\}_{i=1}^N$, trigger patch $p$, source label $y_s$, target label $y_t$, poison budget $M \leq N$, optimization steps $R$, retraining factor $T$
        \ENSURE
        \STATE Train surrogate network or ensemble $F(.\ ; \theta)$ on training data $\mathcal{T}$
        \STATE Select $M$ samples with label  $y_t$ from $\mathcal{T}$ with highest gradient norm
        \STATE Randomly initialize perturbations $\delta_{i=1}^M$

        \FOR {$r$ = 1,\ 2,\ ...\ ,\ $R$ optimizations steps}
            \STATE Compute $ \mathcal{A}(\delta, \theta, p, y_t, y_s)$ and update $\delta_{i=1}^M$ with a step of signed Adam
            \IF {\ $r\mod  \lfloor{R/(T+1)}\rfloor  = 0 $ and $r \ne R$}
                \STATE Retrain $F$ on poisoned training data $\{(x_i + \delta_i, y_i)\}_{i=1}^M \cup \{(x_i, y_i)\}_{i=M+1}^N $ and  update $\theta$
        \ENDIF
        \ENDFOR
        \STATE \textbf{return:} poison perturbations $\delta_{i=1}^M$
    \end{algorithmic} 
    \label{alg1}
\end{algorithm}
\section{Experiments}

In this section, we empirically test the proposed Sleeper Agent backdoor attack on multiple datasets, against black-box settings, using an existing benchmark, and against popular defenses. Details regarding the experimental setup can be found in Appendix \ref{app:imp_det}.

\subsection{Baseline Evaluations}
\label{baseline}

Typically, backdoor attacks are considered successful if poisoned models do not suffer from a significant drop in validation accuracy on images without triggers, but they reliably misclassify images from the source class into the target class when a trigger is applied.  We begin by testing our method in the gray-box setting. In the gray-box setting, we use the same architecture but different random initialization for crafting poisons and testing.  Table \ref{tab:archs} depicts the performance of Sleeper Agent on CIFAR-10 when perturbing $1\%$ of images in the training set with each perturbation constrained in an $\ell_{\infty}$-norm ball of radius $16/255$.  During poison crafting, the surrogate model undergoes four evenly spaced retraining periods ($T=4$), and we test the effectiveness of each surrogate model architecture at generating poisons for victim models of the same architecture.  In subsequent sections, we will extend these experiments to the black-box setting and to an ensemblized attacker.  We observe in these experiments that the poisoned models indeed achieve very similar validation accuracy to their clean counterparts, yet the application of triggers to source class images causes them to be misclassified into the target class as desired.  In Table \ref{tab:budget}, we observe that Sleeper Agent can even be effective when the attacker is only able to poison a very small percentage of the training set.  Note that the success of backdoor attacks depends greatly on the choice of source and target classes, especially since some classes contain very large objects which may dominate the image, even when a trigger is inserted.  As a result, the variance of attack performance is high since we sample class pairs randomly. The poisoning and victim hyperparameters we use for our experiments can be found in Appendix \ref{app:imp_det}.

\begin{table*}
  \caption{\textbf{Baseline evaluations} on CIFAR-10.  Perturbations have $\ell_{\infty}$-norm bounded above by $16/255$, and poison budget is $1\%$ of training images.  
  }
  \label{tab:archs}
  \small
  \centering
  \scalebox{0.85}{\begin{tabular}{lccc}
    \textbf{Architecture}     
    & \textbf{ResNet-18}  
    & \textbf{MobileNetV2} 
    & \textbf{VGG11} 
    \\
    \midrule
    Clean model val\ (\%)   & $92.31 \ (\pm 0.08)$ & $88.19 \ (\pm 0.05)$ & $89.00 \ (\pm 0.03)$\\
    Poisoned model val\ (\%)& $92.16 \ (\pm 0.05)$  & $88.03\ (\pm 0.05)$ & $88.70\ (\pm 0.04)$\\
    Clean model source val\ (\%) & $92.36\ (\pm 0.93)$ & $88.55\ (\pm 1.64)$ & $90.62\ (\pm 1.23)$\\
    Poisoned model source val\ (\%)         & $91.50\ (\pm 0.88)$ & $87.79\ (\pm 1.60)$  & $89.45\ (\pm 1.19)$\\
    Poisoned model patched source val\ (\%)    & $\textbf{12.96}\ (\pm 5.40)$ & $\textbf{21.09}\ (\pm 5.41)$  & $\textbf{17.97}\ (\pm 4.00)$\\
    Attack Success  Rate\ (\%)        & $\textbf{85.27}\ (\pm 5.90)$ & $\textbf{72.92}\ (\pm 6.09)$ & $\textbf{75.15}\ (\pm 5.40)$\\
    
  \end{tabular}}
\end{table*}

\begin{table*}
  \caption{\textbf{The effect of poison budget.}  Experiments on CIFAR-10 with ResNet-18 models \citep{he2016deep}.  Perturbations have $\ell_{\infty}$-norm $\leq 16/255$.  
  }
  \label{tab:budget}
  \small
  \centering
  \scalebox{0.8}{
  \begin{tabular}{lccccc}
    \textbf{Poison Budget}
    & $\textbf{50  (0.1\%)}$  
    & $\textbf{100  (0.2\%)}$  
    & $\textbf{250  (0.5\%)}$ 
    & $\textbf{400 (0.6\%)}$
    & $\textbf{500  (1\%)}$ 
    \\
    \midrule
    Clean model val\ (\%)    & $92.34\ (\pm 0.05)$  & $92.36\ (\pm 0.04)$ & $92.31\ (\pm 0.04)$ & $92.15\ (\pm 0.08)$ & $92.31\ (\pm 0.08)$\\
    Poisoned model val\ (\%) &  $92.33\ (\pm 0.04)$ & $92.34\ (\pm 0.05)$ & $92.25\ (\pm 0.04)$ &  $92.12 \ (\pm 0.06)$ &  $92.16 \ (\pm 0.05)$\\
    Clean model source val\ (\%) & $93.01 \ (\pm 0.69)$ & $91.08 \ (\pm 0.85)$  & $92.43 \ (\pm 0.74)$  & $92.42 \ (\pm 0.80)$ & $92.36 \ (\pm 0.93)$ \\
    Poisoned model source val\ (\%)   & $93.03\ (\pm 0.67)$ & $90.61\ (\pm 0.86)$ & $91.83\ (\pm 0.75)$ & $91.88\ (\pm 0.79)$ & $91.50\ (\pm 0.88)$ \\
    Poisoned model patched source val\ (\%)    &  $\textbf{61.04}\ (\pm 4.27)$ &  $\textbf{40.07}\ (\pm 5.72)$ &  $\textbf{22.77}\ (\pm 4.77)$  & $\textbf{15.88} \ (\pm 4.91)$ & $\textbf{12.96} \ (\pm 5.40)$\\
    Attack Success  Rate\ (\%)        & $\textbf{24.71}\ (\pm 4.10)$ & $\textbf{49.76}\ (\pm 6.21)$  & $\textbf{72.48}\ (\pm 5.24)$  & $\textbf{81.44}\ (\pm 5.25)$ & $\textbf{85.27}\ (\pm 5.90)$  \\
    
  \end{tabular}}
\end{table*}

\textbf{The benefits of ensembling:}  One simple way we can improve the transferability of our backdoor attack across initializations of the same architecture is to craft our poisons on an ensemble of multiple copies of the same architecture but trained using different initializations and different batch sampling during their training procedures. This behavior has also been observed in \citet{huang2020metapoison, geiping2020witches}.  In Table \ref{tab:ensemble}, we observe that this ensembling strategy indeed can offer significant performance boosts, both with and without retraining.


\begin{table*}
  \caption{\textbf{Ensembles} consisting of copies of the same architecture (ResNet-18).  $S$ denotes the size of the ensemble, and $T$ denotes the retraining factor. Experiments are conducted on CIFAR-10, perturbations have $\ell_{\infty}$-norm bounded by $16/255$, and the attacker can poison $1\%$ of training images.}
  \label{tab:ensemble}
  \small
  \centering
  \scalebox{0.9}{
  \begin{tabular}{cccc}
    
    \textbf{Attack} & \textbf{Clean model val} ($\%$) &  \textbf{Poisoned model val} ($\%$)   & \textbf{Attack Success Rate} ($\%$)     \\
    \midrule
    Sleeper Agent ($S=1$, $T=0$)       &   $92.36\ (\pm 0.05)$            &  $92.08\ (\pm 0.08)$            &   $63.49\ (\pm 6.13)$     \\
    Sleeper Agent ($S=2$, $T=0$)       &   $92.10\ (\pm 0.04)$            & $92.12\ (\pm 0.06)$             &   $64.70\ (\pm 5.65)$     \\
    Sleeper Agent ($S=4$, $T=0$)       &   $92.14\ (\pm 0.03)$            & $	91.98 (\pm 0.05)$          &   $\textbf{74.81} \ (\pm 4.10)$\\
    \hline
    Sleeper Agent ($S=2$, $T=4$)       &   $92.11\ (\pm 0.07)$       
    & $92.08\ (\pm0.13)$               &   $87.40\ (\pm 6.23)$     \\
    Sleeper Agent ($S=4$, $T=4$)       &   $92.17\ (\pm 0.03)$
    & $91.81\ (\pm 0.06)$              &   $\textbf{88.45}\ (\pm6.00)$ \\
  \end{tabular}
  }
\end{table*}

\textbf{The black-box setting:}  Now that we have established the transferability of Sleeper Agent across models of the same architecture, we test on the hard black-box scenario where the victim's architecture is completely unknown to the attacker.  This setting has proven extremely challenging for existing methods \citep{schwarzschild2020just}.  Table \ref{tab:black_box} contains four settings.  In the first row, we simply craft the poisons on a single ResNet-18 and transfer these to other models. Second, we craft poisons on an ensemble consisting of two MobileNet-V2 and two ResNet-34 architectures and transfer to the remaining models. Third, for each architecture, we craft poisons with an ensemble consisting of the other two architectures and test on the remaining one.  The second and third scenarios are ensemblized black-box attacks, and we see that Sleeper Agent is effective. In the last row, we perform the same experiment but with the testing model included in the ensemble, and we observe that a single ensemble can craft poisons that are extremely effective on a range of architectures. We choose ResNet-18, MobileNet-V2, and VGG11 as these are common and contain a wide array of structural diversity \citep{he2016deep, sandler2018mobilenetv2, simonyan2014very}.  Additionally, \citet{guo2020practical} considers the case that the attacker uses a weaker surrogate than the defender's model.  We simulate this case by using a VGG11 surrogate and ResNet-18 target.  We find, with a $1\%$ poison budget on CIFAR-10, that Sleeper Agent achieves an attack success rate of $57.47\%$.

\begin{table*}

  \caption{\textbf{Black-box attacks:} First row: Attacks crafted on a single ResNet-18 and transferred. Second row: attacks crafted on MobileNet-V2 and ResNet-34 and transferred. Third row: attacks crafted on the remaining architectures excluding the victim. The ensemble used in the last row includes the victim architecture.  Experiments are conducted on CIFAR-10 and perturbations have $\ell_{\infty}$-norm bounded above by $16/255$, and the attacker can poison $1\%$ of training images.}
  \label{tab:black_box}
  \small
  \centering
  \scalebox{0.9}{
  \begin{tabular}{lccccc}
    \textbf{Attack}     & \textbf{ResNet-18} & \textbf{MobileNet-V2} & \textbf{VGG11} & \textbf{Average}    \\
    \midrule
    Sleeper Agent ($S=1$, $T=4$, ResNet-18)      &   $-$  &   $29.10\%$  &   $31.96\%$  &  $29.86\%$ \\
    Sleeper Agent ($S=4$, $T=0$, MobileNet-V2, ResNet-34)      &   $70.30\%$  &   $-$  &   $46.48\%$  &  $58.44\%$ \\
    Sleeper Agent ($S=4$, $T=0$, victim excluded)     & $63.11\%$ &   $42.40\%$   &  $55.28\%$   &  $53.60\%$ \\
    Sleeper Agent ($S=6$, $T=0$, victim included)    &   $68.46\%$  &  $67.28\%$   &   $85.37\%$ &   $73.30\%$\\
  \end{tabular}
 }
\end{table*}

\textbf{ImageNet evaluations:}  In addition to CIFAR-10, we perform experiments on ImageNet. Table \ref{tab:imagenet} summarizes the performance of Sleeper Agent on ImageNet where attacks are crafted and tested on ResNet-18 and MobileNetV2 models. Each attacker can only perturb $0.05\%$ of training images, and perturbations are constrained in an $\ell_{\infty}$-norm ball of radius $16/255$ - a bound seen in prior poisoning works on ImageNet \citep{fowl2021preventing, geiping2020witches, saha2019hidden}. To have a strong threat model, we use the retraining factor of two ($T=2$) so that the surrogate model is retrained at two evenly spaced intervals. Figure \ref{fig:vis} contains visualizations of the patched sources and the crafted poisons. The details of models and hyperparameters can be found in Appendix \ref{app:imp_det}. Additional experiments on ImageNet and further visualizations are presented in Appendices \ref{app:add_exp} and \ref{app:vis}.

\begin{table*}
  \caption{\textbf{ImageNet evaluations}. Perturbations have $\ell_{\infty}$-norm bounded above by $16/255$, and the poison budget is $0.05\%$ of training images.  
  }
  \label{tab:imagenet}
  \small
  \centering
  \begin{adjustbox}{width=0.55\columnwidth,center}
  \begin{tabular}{lcc}
    \textbf{Architecture}     
    & \textbf{ResNet-18}  
    & \textbf{MobileNetV2} 
    \\ 
    \midrule
    Clean model val\ (\%)   & $69.76$ & $71.88$ \\
    Poisoned model val\ (\%)& $67.84 \ (\pm 0.10)$  & $68.60\ (\pm 0.03)$ \\
    Attack Success  Rate\ (\%)        & $\textbf{44.00}\ (\pm 6.73)$ & $\textbf{41.00}\ (\pm 3.31)$\\
  \end{tabular}
  \end{adjustbox}
\end{table*}

\subsection{Comparison to Other Methods}
\label{sec:comp_others}

There are several existing clean-label hidden-trigger backdoor attacks that claim success in settings different than ours. In order to further demonstrate the success of our method, we compare our poisons to ones generated from these methods in our strict threat model of from-scratch training. In these experiments, poisons are generated by our attack, clean label backdoor, and hidden trigger backdoor. All poison trials have the same randomly selected source-target class pairs, the same budget, and the same $\varepsilon$-bound (Note: clean-label backdoor originally did not use $\ell_\infty$ bounds, so we adjust the opacity of their perturbations to ensure the constraint is satisfied). We then train a randomly initialized network from scratch on these poisons and evaluate success over $1000$ patched source images. We test three popular architectures and find that our attack significantly outperforms both methods and is the only backdoor method to exceed single digit success rates, confirming the findings of \citet{schwarzschild2020just} on the fragility of these existing methods. See Table \ref{tab:benchmark} for full results. 


\begin{table*}
  \caption{\textbf{Benchmark results on CIFAR-10}. Comparison of our method to popular ``clean-label'' attacks. Results averaged over the same source/target pairs with $\epsilon = 16/255$ and poison budget $1\%$.}
  \label{tab:benchmark}
  \small
  \centering
  \begin{adjustbox}{width=0.9\columnwidth,center}
  \begin{tabular}{ccccc}
    \textbf{Attack}     & \textbf{ResNet-18} & \textbf{MobileNetV2} & \textbf{VGG11} & \textbf{Average}    \\
    \midrule
    Hidden-Trigger Backdoor \citep{saha2019hidden}     &   $3.50 \%$   &  $3.76 \%$   &  $5.02 \%$  &   $4.09 \%$\\
    Clean-Label Backdoor \citep{turner2019label}    &  $2.78 \%$   &   $3.50\%$  &  $4.70\%$  & $3.66\%$  \\
    Sleeper Agent (Ours)    &  $\textbf{78.84}\%$    &    $\textbf{75.96}\%$ &  $\textbf{86.60}\%$ &  $\textbf{80.47}\%$ \\
  \end{tabular}
  \end{adjustbox}
\end{table*}

\subsection{Defenses}
\label{subsec:defenses}

A selling point for hidden trigger backdoor attacks is that the trigger that is used to induce misclassification at test-time is not present in any training data, thus making inspection based defenses, or automated pattern matching more difficult. However, there exist numerous defenses, aside from visual inspection, that have been proposed to mitigate the effects of poisoning - both backdoor and other attacks. We test our method against a number of popular defenses.

\textbf{Spectral Signatures}: This defense, proposed in \citet{tran2018spectral}, aims to filter a pre-selected amount of training data based upon correlations with singular vectors of the feature covariance matrix. This defense was originally intended to detect triggers used in backdoor attacks.

\textbf{Activation Clustering}: \citet{chen2018detecting} clusters activation patterns to detect anomalous inputs. Unlike the spectral signatures defense, this defense does not filter a pre-selected volume of data.

\textbf{DPSGD}: Poison defenses based on differentially private SGD \citep{abadi2016deep} have also been proposed \citep{hong2020effectiveness}. Differentially private learning inures models to small changes in training data, which provably imbues robustness to poisoned data.

\textbf{Data Augmentations}: Recent work has suggested that strong data augmentations, such as mixup, break data poisoning \citep{borgnia2021dp}. This has been confirmed in recent benchmark tests which demonstrate many poisoning techniques are brittle to slight changes in victim training routine \citep{schwarzschild2020just}. We test against mixup augmentation \citep{zhang2017mixup}.

\textbf{STRIP}: \citet{gao2019strip} proposes to add strong perturbations by superimposing input images at test time to detect the backdoored inputs based on the entropy of the predicted class distribution. If the entropy is lower than a predefined threshold, the input is considered backdoored and is rejected.

\textbf{NeuralCleanse}: \citet{wang2019neural} proposes a defense designed for traditional backdoor attacks by reconstructing the maximally adversarial trigger used to backdoor a model. While this defense was not designed for hidden trigger backdoor attacks, we experiment with this as a \emph{detection} defense wherein we test whether NeuralCleanse can detect the backdoored class. This modification is denoted by NeuralCleanse*. In our trials, NeuralCleanse* does not successfully detect any of the backdoored classes - as determined by taking the maximum mask MAD (see \citet{wang2019neural}). Neural Cleanse does not produce an anomaly score $>2$ (their characterization of detecting outliers) for the backdoored class in \emph{any} of our experiments. 

We find that across the board, all of these defenses exhibit a robustness-accuracy trade-off.  Many of these defenses do not reliably nullify the attack, and defenses that do degrade attack success also induce such a large drop in validation accuracy that they are unattractive options for practitioners. For example, to lower the attack success to an average of $13.14\%$, training with DPSGD degrades natural accuracy on CIFAR-10 to $70\%$. See Table \ref{tab:defenses} for the complete results of these experiments. Additional evaluations on recent defenses are presented in Appendix \ref{app:add_def}.

\begin{table*}
  \caption{\textbf{Defenses}. Experiments are conducted on CIFAR-10 with ResNet-18 models, perturbations have $\ell_{\infty}$-norm bounded above by $16/255$, and poison budget is $1\%$ of training images.}
  
  \label{tab:defenses}
  \centering
  \small
  \begin{adjustbox}{width=0.7\columnwidth,center}
  \begin{tabular}{ccc}
    \textbf{Defense}     & \textbf{Attack Success Rate} ($\%$)  & \textbf{Validation Accuracy} ($\%$) \\
    \midrule
    Spectral Signatures & $37.17 \ (\pm 10.10)$ & $89.94 \ (\pm 0.19)$\\
    Activation Clustering & $15.17 \ (\pm 5.38)$ & $72.38 \ (\pm 0.48)$\\ 
    DPSGD & $13.14 \ (\pm 4.49) $ & $70.00 \ (\pm 0.17)$\\ 
    Data Augmentation & $69.75 \ (\pm 10.77)$ & $91.32 \ (\pm 0.12)$\\ 
    STRIP & $62.68 \ (\pm 4.90)$ & $92.23 \ (\pm 0.05)$\\ 
    NeuralCleanse* &  $85.27  \ (\pm 5.90)$ & $92.31 \ (\pm 0.08)$\\
  \end{tabular}
  \end{adjustbox}
\end{table*}

\subsection{Sleeper Agent Can Poison Images in Any Class}

Typical backdoor attacks which rely on label flips or feature collisions can only function when poisons come from the source and/or target classes \citep{saha2019hidden, turner2019label}.  This restriction may be a serious limitation in practice.  In contrast, we show that Sleeper Agent can be effective even when we poison images drawn from all classes.  To take advantage of our data selection strategy, we select poisons with maximum gradient norm across all classes. Table \ref{tab:random_poison} contains the performance of Sleeper Agent in the aforementioned setting.

\begin{table*}
  \caption{\textbf{Random poisons}. Experiments are conducted on CIFAR-10 with ResNet-18 models.  Perturbations have $\ell_{\infty}$-norm bounded above by $16/255$ and poisons are drawn from all classes.  
  }
  \label{tab:random_poison}
  \centering
  \small
  \begin{adjustbox}{width=0.70\columnwidth,center}
  \begin{tabular}{ccc}
    \textbf{Attack}  & \textbf{Poison budget}   & \textbf{Attack Success Rate} ($\%$)   \\
    \midrule
    Sleeper Agent (S = 1, T = 4)   &  $1 \%$   &  $\textbf{41.90} \ (\pm 7.16)$ \\
    Sleeper Agent (S = 1, T = 4)   &  $3 \%$   &  $\textbf{66.51} \ (\pm 6.90)$ \\
    
  \end{tabular}
  \end{adjustbox}
\end{table*}

\subsection{Evaluations Under Hard  \texorpdfstring{$\ell_{\infty}$}{linf}-norm Constraints}

While existing works on backdoor attacks consider poisons with $\ell_{\infty}$-norm bounded above by $16/255$ as an imperceptible threat \citep{saha2019hidden,turner2019label}, \citet{nguyen2021wanet} shows that human inspection can detect poisoned samples effectively. This inspection might mitigate the threat of large perturbations. To bypass this possibility, we conduct our baseline experiments on CIFAR-10 using perturbations with small $\ell_{\infty}$-norms. From Table \ref{tab:norm}, we observe that our threat model is effective even with an $\ell_{\infty}$-norm bounded above by $8/255$. Visualizations can be found in Appendix \ref{app:vis}.

\begin{table*}
  \caption{\textbf{Evaluation under different $\ell_{\infty}$-norm}. Experiments are conducted on CIFAR-10 with ResNet-18 models, and the poison budget is $1\%$ of training images.}
  
  \label{tab:norm}
  \small
  \centering
  \begin{adjustbox}{width=0.5\columnwidth,center}
  \begin{tabular}{cc}
    \textbf{Perturbation $\ell_{\infty}$-norm} & $\textbf{Attack Success Rate\ (\%)\ }$ \\
    \midrule
    $8/255$ & $37.32 \ (\pm 8.33)$ \\
    $10/255$ & $55.75 \ (\pm 8.12)$ \\
    $12/255$ & $63.31 \ (\pm 8.84)$ \\
    $14/255$ & $78.03 \ (\pm 7.13)$ \\
    $16/255$ & $85.27 \ (\pm 5.90)$\\
  \end{tabular}
  \end{adjustbox}
\end{table*}

\subsection{Ablation Studies}

Here, we analyze the importance of each technique in our algorithm via ablation studies. We focus on three aspects of our method: 1) patch location, 2) retraining during poison crafting, 3) poison selection, and 4) retraining factor. Table \ref{tab:ablation_cifar} details the combinations and their effects on poison success. We find that randomizing patch location improves poisoning success, and both retraining and data selection based on maximum gradient significantly improve poison performance. Combining all three boosts poison success more than four-fold. To further show the importance of retraining, we conduct more experiments with and without retraining on ImageNet. From Table \ref{tab:imagenet_ablation}, we infer that retraining is essential. Additional ablations studies are found in Appendix \ref{app:add_exp}.

\begin{table*}[ht]
  \caption{\textbf{CIFAR-10 ablation studies.} Investigation of the effects of random patch-location, retraining, and data selection. Experiments are conducted on CIFAR-10 with ResNet-18 models, perturbations have $\ell_{\infty}$-norm bounded above by $16/255$, and poison budget is $1\%$ of training images.}
  \label{tab:ablation_cifar}
  \small
  \centering
  \begin{adjustbox}{width=0.75\columnwidth,center}
  \begin{tabular}{lc}
    \textbf{Attack setup}     & \textbf{Attack Success Rate} ($\%$)    \\
    \midrule
    Fix patch-location (bottom-right corner)      &   $19.25\ (\pm 3.01) $    \\
    Random patch-location   &   $33.95 \ (\pm 4.57)$   \\
    Random patch-location + retraining     & $59.42 \ (\pm 5.78)$         \\
    Random patch-location + data selection     & $63.49 \ (\pm 6.13)$ \\
    Random patch-location + retraining + data selection & $\textbf{85.27} \ (\pm 5.90)$  
  \end{tabular}
  \end{adjustbox}
\end{table*}

\begin{table*}[h!]
  \caption{\textbf{ImageNet ablation studies}.  Perturbations have $\ell_{\infty}$-norm bounded above by $16/255$, and the poison budget is $0.05\%$ of training images.}
  \label{tab:imagenet_ablation}
  \small
  \centering
  \begin{adjustbox}{width=0.5\columnwidth,center}
  \begin{tabular}{cc}
    \textbf{Attack} & \textbf{Attack Success Rate}  ($\%$)   \\
    \midrule
    Sleeper Agent\ (S = 1, T = 0) &  $22.00 \ (\pm 5.65)$ \\
    Sleeper Agent\ (S = 1, T = 2) &   $\textbf{44.00} \ (\pm 6.73)$
  \end{tabular}
  \end{adjustbox}
\end{table*}

\section{Broader Impact and Limitations}
\label{sec:br_lim}

In this work, we illuminate a new scalable backdoor attack that could be used to stealthily compromise security-critical systems.  We hope that by highlighting the potential danger of this nefarious threat model, our work will give rise to stronger defenses and will encourage caution on the part of practitioners.

While on average, our method is effective, the variance is large, and the success of our method can range from almost all patched images being misclassified to low success. This behavior has previously been observed in \cite{schwarzschild2020just}.  In real-world scenarios, datasets are often noisy and imbalanced, so training behavior may be mysterious.  As a result, practitioners should be cautious in their expectations that methods developed on datasets like CIFAR-10 and ImageNet will work on their own problems.

\section{Conclusion}

In this work, we present the first hidden-trigger backdoor attack that is effective against deep networks trained from scratch.  This is a challenging setting for backdoor attacks, and existing attacks typically operate in less strict settings.  Nonetheless, we choose the strict setting because practitioners often train networks from scratch in real-world applications, and patched poisons may be easily visible upon human inspection.  In order to accomplish the above goal, we use a gradient matching objective as a surrogate for the bilevel optimization problem, and we add features such as re-training and data selection in order to significantly enhance the performance of our method, Sleeper Agent.


\section*{Acknowledgements}
This work was supported by DARPA GARD under contracts \#HR00112020007 and HR001119S0026-GARD-FP-052, the DARPA YFA program, the ONR MURI Program under the Grant N00014-20-1-2787, and the National Science Foundation DMS program. Further support was provided by JP Morgan Chase and Capital One Bank.

\bibliography{main}

\begin{thebibliography}{44}
\providecommand{\natexlab}[1]{#1}
\providecommand{\url}[1]{\texttt{#1}}
\expandafter\ifx\csname urlstyle\endcsname\relax
  \providecommand{\doi}[1]{doi: #1}\else
  \providecommand{\doi}{doi: \begingroup \urlstyle{rm}\Url}\fi

\bibitem[Abadi et~al.(2016)Abadi, Chu, Goodfellow, McMahan, Mironov, Talwar,
  and Zhang]{abadi2016deep}
Martin Abadi, Andy Chu, Ian Goodfellow, H~Brendan McMahan, Ilya Mironov, Kunal
  Talwar, and Li~Zhang.
\newblock Deep learning with differential privacy.
\newblock In \emph{Proceedings of the 2016 ACM SIGSAC conference on computer
  and communications security}, pages 308--318, 2016.

\bibitem[Bagdasaryan et~al.(2020)Bagdasaryan, Veit, Hua, Estrin, and
  Shmatikov]{bagdasaryan2020backdoor}
Eugene Bagdasaryan, Andreas Veit, Yiqing Hua, Deborah Estrin, and Vitaly
  Shmatikov.
\newblock How to backdoor federated learning.
\newblock In \emph{International Conference on Artificial Intelligence and
  Statistics}, pages 2938--2948. PMLR, 2020.

\bibitem[Barni et~al.(2019)Barni, Kallas, and Tondi]{barni2019new}
Mauro Barni, Kassem Kallas, and Benedetta Tondi.
\newblock A new backdoor attack in cnns by training set corruption without
  label poisoning.
\newblock In \emph{2019 IEEE International Conference on Image Processing
  (ICIP)}, pages 101--105. IEEE, 2019.

\bibitem[Biggio et~al.(2012)Biggio, Nelson, and Laskov]{biggio2012poisoning}
Battista Biggio, Blaine Nelson, and Pavel Laskov.
\newblock Poisoning attacks against support vector machines.
\newblock In \emph{Proceedings of the 29th International Coference on
  International Conference on Machine Learning}, pages 1467--1474, 2012.

\bibitem[Borgnia et~al.(2021)Borgnia, Geiping, Cherepanova, Fowl, Gupta,
  Ghiasi, Huang, Goldblum, and Goldstein]{borgnia2021dp}
Eitan Borgnia, Jonas Geiping, Valeriia Cherepanova, Liam Fowl, Arjun Gupta,
  Amin Ghiasi, Furong Huang, Micah Goldblum, and Tom Goldstein.
\newblock Dp-instahide: Provably defusing poisoning and backdoor attacks with
  differentially private data augmentations.
\newblock \emph{arXiv preprint arXiv:2103.02079}, 2021.

\bibitem[Chen et~al.(2019)Chen, Carvalho, Baracaldo, Ludwig, Edwards, Lee,
  Molloy, and Srivastava]{chen2018detecting}
Bryant Chen, Wilka Carvalho, Nathalie Baracaldo, Heiko Ludwig, Benjamin
  Edwards, Taesung Lee, Ian Molloy, and Biplav Srivastava.
\newblock Detecting backdoor attacks on deep neural networks by activation
  clustering.
\newblock In \emph{SafeAI@ AAAI}, 2019.

\bibitem[Chen et~al.(2017)Chen, Liu, Li, Lu, and Song]{chen2017targeted}
Xinyun Chen, Chang Liu, Bo~Li, Kimberly Lu, and Dawn Song.
\newblock Targeted backdoor attacks on deep learning systems using data
  poisoning.
\newblock \emph{arXiv preprint arXiv:1712.05526}, 2017.

\bibitem[Feng et~al.(2019)Feng, Cai, and Zhou]{feng2019learning}
Ji~Feng, Qi-Zhi Cai, and Zhi-Hua Zhou.
\newblock Learning to confuse: generating training time adversarial data with
  auto-encoder.
\newblock \emph{Advances in Neural Information Processing Systems}, 32, 2019.

\bibitem[Fowl et~al.(2021{\natexlab{a}})Fowl, Chiang, Goldblum, Geiping,
  Bansal, Czaja, and Goldstein]{fowl2021preventing}
Liam Fowl, Ping-yeh Chiang, Micah Goldblum, Jonas Geiping, Arpit Bansal, Wojtek
  Czaja, and Tom Goldstein.
\newblock Preventing unauthorized use of proprietary data: Poisoning for secure
  dataset release.
\newblock \emph{arXiv preprint arXiv:2103.02683}, 2021{\natexlab{a}}.

\bibitem[Fowl et~al.(2021{\natexlab{b}})Fowl, Goldblum, Chiang, Geiping, Czaja,
  and Goldstein]{fowl2021adversarial}
Liam Fowl, Micah Goldblum, Ping-yeh Chiang, Jonas Geiping, Wojciech Czaja, and
  Tom Goldstein.
\newblock Adversarial examples make strong poisons.
\newblock \emph{Advances in Neural Information Processing Systems},
  34:\penalty0 30339--30351, 2021{\natexlab{b}}.

\bibitem[Gao et~al.(2019)Gao, Xu, Wang, Chen, Ranasinghe, and
  Nepal]{gao2019strip}
Yansong Gao, Change Xu, Derui Wang, Shiping Chen, Damith~C Ranasinghe, and
  Surya Nepal.
\newblock Strip: A defence against trojan attacks on deep neural networks.
\newblock In \emph{Proceedings of the 35th Annual Computer Security
  Applications Conference}, pages 113--125, 2019.

\bibitem[Geiping et~al.(2021)Geiping, Fowl, Huang, Czaja, Taylor, Moeller, and
  Goldstein]{geiping2020witches}
Jonas Geiping, Liam Fowl, W~Ronny Huang, Wojciech Czaja, Gavin Taylor, Michael
  Moeller, and Tom Goldstein.
\newblock Witches' brew: Industrial scale data poisoning via gradient matching.
\newblock \emph{ICLR}, 2021.

\bibitem[Goldblum et~al.(2022)Goldblum, Tsipras, Xie, Chen, Schwarzschild,
  Song, Madry, Li, and Goldstein]{goldblum2020data}
Micah Goldblum, Dimitris Tsipras, Chulin Xie, Xinyun Chen, Avi Schwarzschild,
  Dawn Song, Aleksander Madry, Bo~Li, and Tom Goldstein.
\newblock Dataset security for machine learning: Data poisoning, backdoor
  attacks, and defenses.
\newblock \emph{IEEE Transactions on Pattern Analysis and Machine
  Intelligence}, 2022.

\bibitem[Gu et~al.(2017)Gu, Dolan-Gavitt, and Garg]{gu2017badnets}
Tianyu Gu, Brendan Dolan-Gavitt, and Siddharth Garg.
\newblock Badnets: Identifying vulnerabilities in the machine learning model
  supply chain.
\newblock \emph{arXiv preprint arXiv:1708.06733}, 2017.

\bibitem[Guo and Liu(2020)]{guo2020practical}
Junfeng Guo and Cong Liu.
\newblock Practical poisoning attacks on neural networks.
\newblock In \emph{European Conference on Computer Vision}, pages 142--158.
  Springer, 2020.

\bibitem[He et~al.(2016)He, Zhang, Ren, and Sun]{he2016deep}
Kaiming He, Xiangyu Zhang, Shaoqing Ren, and Jian Sun.
\newblock Deep residual learning for image recognition.
\newblock In \emph{Proceedings of the IEEE conference on computer vision and
  pattern recognition}, pages 770--778, 2016.

\bibitem[Hong et~al.(2020)Hong, Chandrasekaran, Kaya, Dumitra{\c{s}}, and
  Papernot]{hong2020effectiveness}
Sanghyun Hong, Varun Chandrasekaran, Yi{\u{g}}itcan Kaya, Tudor Dumitra{\c{s}},
  and Nicolas Papernot.
\newblock On the effectiveness of mitigating data poisoning attacks with
  gradient shaping.
\newblock \emph{arXiv preprint arXiv:2002.11497}, 2020.

\bibitem[Huang et~al.(2020{\natexlab{a}})Huang, Ma, Erfani, Bailey, and
  Wang]{huang2021unlearnable}
Hanxun Huang, Xingjun Ma, Sarah~Monazam Erfani, James Bailey, and Yisen Wang.
\newblock Unlearnable examples: Making personal data unexploitable.
\newblock In \emph{International Conference on Learning Representations},
  2020{\natexlab{a}}.

\bibitem[Huang et~al.(2020{\natexlab{b}})Huang, Geiping, Fowl, Taylor, and
  Goldstein]{huang2020metapoison}
W~Ronny Huang, Jonas Geiping, Liam Fowl, Gavin Taylor, and Tom Goldstein.
\newblock Metapoison: Practical general-purpose clean-label data poisoning.
\newblock \emph{Advances in Neural Information Processing Systems},
  33:\penalty0 12080--12091, 2020{\natexlab{b}}.

\bibitem[Li et~al.(2019)Li, Zhao, Yu, Xue, Kaafar, and Zhu]{li2019invisible}
Shaofeng Li, Benjamin Zi~Hao Zhao, Jiahao Yu, Minhui Xue, Dali Kaafar, and
  Haojin Zhu.
\newblock Invisible backdoor attacks against deep neural networks.
\newblock \emph{arXiv preprint arXiv:1909.02742}, 2019.

\bibitem[Li et~al.(2020)Li, Xue, Zhao, Zhu, and Zhang]{li2020invisible}
Shaofeng Li, Minhui Xue, Benjamin Zhao, Haojin Zhu, and Xinpeng Zhang.
\newblock Invisible backdoor attacks on deep neural networks via steganography
  and regularization.
\newblock \emph{IEEE Transactions on Dependable and Secure Computing}, 2020.

\bibitem[Li et~al.(2021{\natexlab{a}})Li, Lyu, Koren, Lyu, Li, and
  Ma]{li2021anti}
Yige Li, Xixiang Lyu, Nodens Koren, Lingjuan Lyu, Bo~Li, and Xingjun Ma.
\newblock Anti-backdoor learning: Training clean models on poisoned data.
\newblock \emph{Advances in Neural Information Processing Systems},
  34:\penalty0 14900--14912, 2021{\natexlab{a}}.

\bibitem[Li et~al.(2022)Li, Jiang, Li, and Xia]{li2020backdoor}
Yiming Li, Yong Jiang, Zhifeng Li, and Shu-Tao Xia.
\newblock Backdoor learning: A survey.
\newblock \emph{IEEE Transactions on Neural Networks and Learning Systems},
  2022.

\bibitem[Li et~al.(2021{\natexlab{b}})Li, Li, Wu, Li, He, and
  Lyu]{li2021invisible}
Yuezun Li, Yiming Li, Baoyuan Wu, Longkang Li, Ran He, and Siwei Lyu.
\newblock Invisible backdoor attack with sample-specific triggers.
\newblock In \emph{Proceedings of the IEEE/CVF International Conference on
  Computer Vision}, pages 16463--16472, 2021{\natexlab{b}}.

\bibitem[Liu et~al.(2017)Liu, Ma, Aafer, Lee, Zhai, Wang, and
  Zhang]{liu2017trojaning}
Yingqi Liu, Shiqing Ma, Yousra Aafer, Wen-Chuan Lee, Juan Zhai, Weihang Wang,
  and Xiangyu Zhang.
\newblock Trojaning attack on neural networks.
\newblock 2017.

\bibitem[Liu et~al.(2020)Liu, Ma, Bailey, and Lu]{liu2020reflection}
Yunfei Liu, Xingjun Ma, James Bailey, and Feng Lu.
\newblock Reflection backdoor: A natural backdoor attack on deep neural
  networks.
\newblock In \emph{European Conference on Computer Vision}, pages 182--199.
  Springer, 2020.

\bibitem[{Muñoz-González} et~al.(2017){Muñoz-González}, Biggio, Demontis,
  Paudice, Wongrassamee, Lupu, and Roli]{munoz-gonzalez_towards_2017}
Luis {Muñoz-González}, Battista Biggio, Ambra Demontis, Andrea Paudice, Vasin
  Wongrassamee, Emil~C. Lupu, and Fabio Roli.
\newblock Towards {{Poisoning}} of {{Deep Learning Algorithms}} with
  {{Back}}-gradient {{Optimization}}.
\newblock In \emph{Proceedings of the 10th {{ACM Workshop}} on {{Artificial
  Intelligence}} and {{Security}}}, {{AISec}} '17, pages 27--38, {New York, NY,
  USA}, 2017. {ACM}.
\newblock ISBN 978-1-4503-5202-4.
\newblock \doi{10.1145/3128572.3140451}.

\bibitem[Nguyen and Tran(2020)]{nguyen2021wanet}
Tuan~Anh Nguyen and Anh~Tuan Tran.
\newblock Wanet-imperceptible warping-based backdoor attack.
\newblock In \emph{International Conference on Learning Representations}, 2020.

\bibitem[Russakovsky et~al.(2015)Russakovsky, Deng, Su, Krause, Satheesh, Ma,
  Huang, Karpathy, Khosla, Bernstein, et~al.]{russakovsky2015imagenet}
Olga Russakovsky, Jia Deng, Hao Su, Jonathan Krause, Sanjeev Satheesh, Sean Ma,
  Zhiheng Huang, Andrej Karpathy, Aditya Khosla, Michael Bernstein, et~al.
\newblock Imagenet large scale visual recognition challenge.
\newblock \emph{International journal of computer vision}, 115\penalty0
  (3):\penalty0 211--252, 2015.

\bibitem[Saha et~al.(2020)Saha, Subramanya, and Pirsiavash]{saha2019hidden}
Aniruddha Saha, Akshayvarun Subramanya, and Hamed Pirsiavash.
\newblock Hidden trigger backdoor attacks.
\newblock In \emph{Proceedings of the AAAI conference on artificial
  intelligence}, volume~34, pages 11957--11965, 2020.

\bibitem[Sandler et~al.(2018)Sandler, Howard, Zhu, Zhmoginov, and
  Chen]{sandler2018mobilenetv2}
Mark Sandler, Andrew Howard, Menglong Zhu, Andrey Zhmoginov, and Liang-Chieh
  Chen.
\newblock Mobilenetv2: Inverted residuals and linear bottlenecks.
\newblock In \emph{Proceedings of the IEEE conference on computer vision and
  pattern recognition}, pages 4510--4520, 2018.

\bibitem[Schwarzschild et~al.(2021)Schwarzschild, Goldblum, Gupta, Dickerson,
  and Goldstein]{schwarzschild2020just}
Avi Schwarzschild, Micah Goldblum, Arjun Gupta, John~P Dickerson, and Tom
  Goldstein.
\newblock Just how toxic is data poisoning? a unified benchmark for backdoor
  and data poisoning attacks.
\newblock In \emph{International Conference on Machine Learning}, pages
  9389--9398. PMLR, 2021.

\bibitem[Shafahi et~al.(2018)Shafahi, Huang, Najibi, Suciu, Studer, Dumitras,
  and Goldstein]{shafahi2018poison}
Ali Shafahi, W~Ronny Huang, Mahyar Najibi, Octavian Suciu, Christoph Studer,
  Tudor Dumitras, and Tom Goldstein.
\newblock Poison frogs! targeted clean-label poisoning attacks on neural
  networks.
\newblock \emph{Advances in neural information processing systems}, 31, 2018.

\bibitem[Shen et~al.(2019)Shen, Zhu, and Ma]{shen2019tensorclog}
Juncheng Shen, Xiaolei Zhu, and De~Ma.
\newblock Tensorclog: An imperceptible poisoning attack on deep neural network
  applications.
\newblock \emph{IEEE Access}, 7:\penalty0 41498--41506, 2019.

\bibitem[Simonyan and Zisserman(2014)]{simonyan2014very}
Karen Simonyan and Andrew Zisserman.
\newblock Very deep convolutional networks for large-scale image recognition.
\newblock \emph{arXiv preprint arXiv:1409.1556}, 2014.

\bibitem[Steinhardt et~al.(2017)Steinhardt, Koh, and
  Liang]{steinhardt_certified_2017}
Jacob Steinhardt, Pang Wei~W Koh, and Percy~S Liang.
\newblock Certified {{Defenses}} for {{Data Poisoning Attacks}}.
\newblock In \emph{Advances in {{Neural Information Processing Systems}} 30},
  pages 3517--3529. {Curran Associates, Inc.}, 2017.

\bibitem[Tran et~al.(2018)Tran, Li, and Madry]{tran2018spectral}
Brandon Tran, Jerry Li, and Aleksander Madry.
\newblock Spectral signatures in backdoor attacks.
\newblock \emph{Advances in neural information processing systems}, 31, 2018.

\bibitem[Turner et~al.(2019)Turner, Tsipras, and Madry]{turner2019label}
Alexander Turner, Dimitris Tsipras, and Aleksander Madry.
\newblock Label-consistent backdoor attacks.
\newblock \emph{arXiv preprint arXiv:1912.02771}, 2019.

\bibitem[Wang et~al.(2019)Wang, Yao, Shan, Li, Viswanath, Zheng, and
  Zhao]{wang2019neural}
Bolun Wang, Yuanshun Yao, Shawn Shan, Huiying Li, Bimal Viswanath, Haitao
  Zheng, and Ben~Y Zhao.
\newblock Neural cleanse: Identifying and mitigating backdoor attacks in neural
  networks.
\newblock In \emph{2019 IEEE Symposium on Security and Privacy (SP)}, pages
  707--723. IEEE, 2019.

\bibitem[Wenger et~al.(2021)Wenger, Passananti, Bhagoji, Yao, Zheng, and
  Zhao]{wenger2021backdoor}
Emily Wenger, Josephine Passananti, Arjun~Nitin Bhagoji, Yuanshun Yao, Haitao
  Zheng, and Ben~Y Zhao.
\newblock Backdoor attacks against deep learning systems in the physical world.
\newblock In \emph{Proceedings of the IEEE/CVF Conference on Computer Vision
  and Pattern Recognition}, pages 6206--6215, 2021.

\bibitem[Wu and Wang(2021)]{wu2021adversarial}
Dongxian Wu and Yisen Wang.
\newblock Adversarial neuron pruning purifies backdoored deep models.
\newblock \emph{Advances in Neural Information Processing Systems},
  34:\penalty0 16913--16925, 2021.

\bibitem[Zhang et~al.(2018)Zhang, Cisse, Dauphin, and
  Lopez-Paz]{zhang2017mixup}
Hongyi Zhang, Moustapha Cisse, Yann~N Dauphin, and David Lopez-Paz.
\newblock mixup: Beyond empirical risk minimization.
\newblock In \emph{International Conference on Learning Representations}, 2018.

\bibitem[Zhao et~al.(2021)Zhao, Mopuri, and Bilen]{zhao2020dataset}
Bo~Zhao, Konda~Reddy Mopuri, and Hakan Bilen.
\newblock Dataset condensation with gradient matching.
\newblock In \emph{Ninth International Conference on Learning Representations
  2021}, 2021.

\bibitem[Zhu et~al.(2019)Zhu, Huang, Li, Taylor, Studer, and
  Goldstein]{zhu2019transferable}
Chen Zhu, W~Ronny Huang, Hengduo Li, Gavin Taylor, Christoph Studer, and Tom
  Goldstein.
\newblock Transferable clean-label poisoning attacks on deep neural nets.
\newblock In \emph{International Conference on Machine Learning}, pages
  7614--7623. PMLR, 2019.

\end{thebibliography}
\bibliographystyle{plainnat}



\newpage


\appendix


  


\section{Additional Experiments}
\label{app:add_exp}


\subsection{Gradient Alignment Throughout Training}
\label{subsec:alignment}

\begin{figure}[ht]
    \centering
    \includegraphics[width=0.65\linewidth]{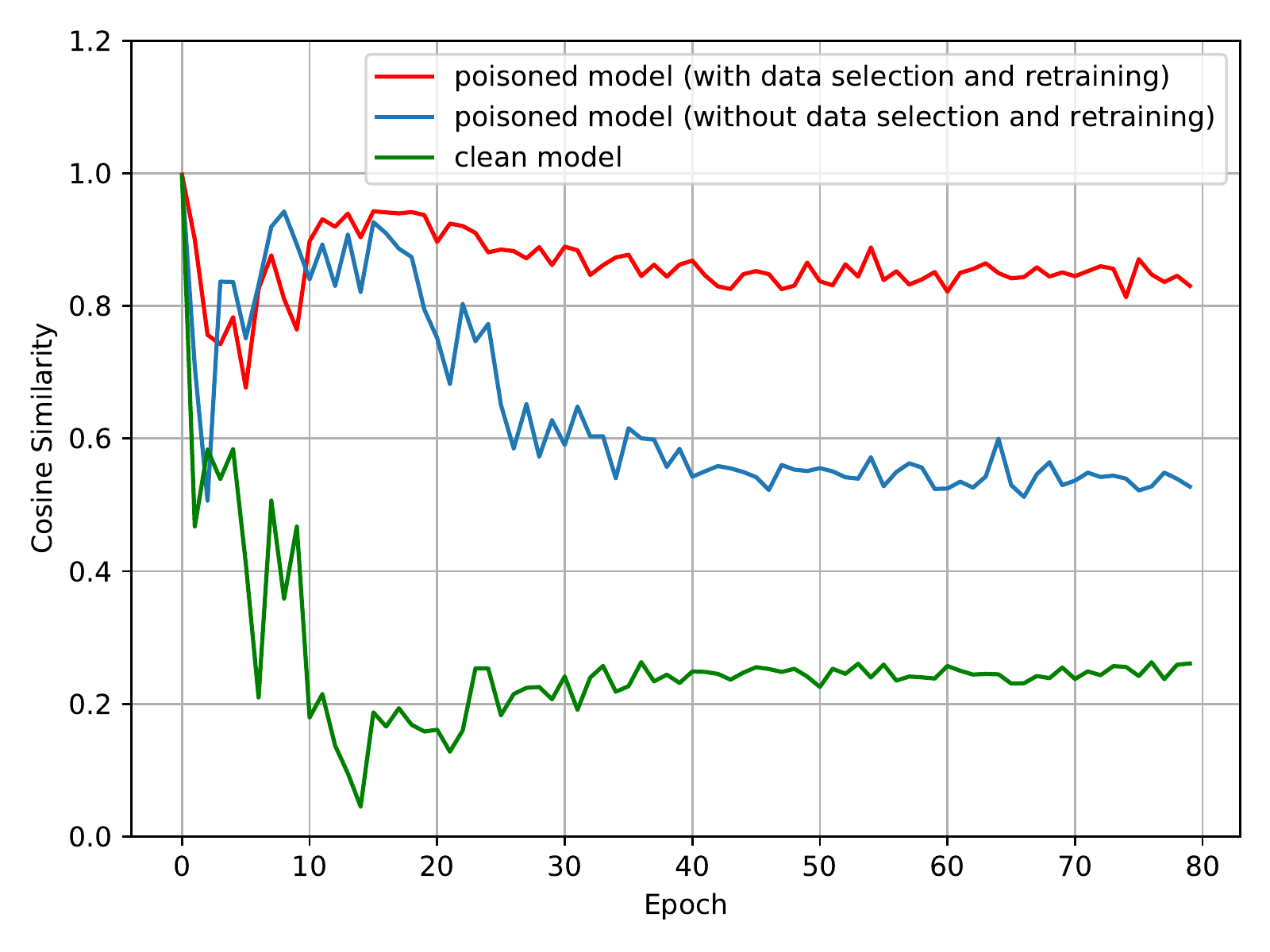}
    \caption{Cosine Similarity, per epoch, between the adversarial gradient $\nabla_\theta \mathcal{L}_{adv}$ and gradient of the poison examples (clean examples from the target class in the case of clean model training) $\nabla_\theta \mathcal{L}_{train}$  for two different poisoned models and a clean model. Experiments are conducted on CIFAR-10 with ResNet-18 models.} 
    \label{fig:cosine}
\end{figure}

In order to demonstrate that the gradients of the poison examples are well aligned with the adversarial gradient throughout the training of the victim model, we visualize the cosine similarity between the adversarial gradient and the poison examples in multiple settings across epochs of training. Figure \ref{fig:cosine} contains three experiments. First, we train a clean model where the attack's success rate is very low (almost zero). Second, we train a poisoned model without data selection or retraining. And third, we employ poisons that have been generated utilizing data selection and retraining techniques. As shown in Table \ref{tab:ablation_cifar}, the average attack success rate for the second and third experiments is $33.95\%$ and $85.27\%$, respectively. Figure \ref{fig:cosine} shows that a successful attack yields far superior gradient alignment and hence a high attack success rate. In addition, these experiments demonstrate that gradient alignment, data selection, and retraining all work together collaboratively.

\subsection{Patch Choice}
\label{app:patch_choice}

\begin{figure}[h]
    \centering
    \includegraphics[width=0.5\linewidth]{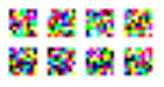}
    \caption{Sample random patches}
    \label{fig:random_patch}
\end{figure}

Sleeper Agent is designed in a way that the backdoor attack is efficient for any random patch the threat model uses for crafting poisons. To show this, we conduct the same baseline experiments discussed in Section \ref{baseline} using different random patches that are generated using a Bernoulli distribution. From Table \ref{tab:bernoulli}, we observe that the choice of the patch does not affect Sleeper Agent's success rate. Figure \ref{fig:random_patch} depicts few samples of the random patches we use for the experiments presented in Table \ref{tab:bernoulli}.

\begin{table*}[h]
  \caption{\textbf{Baseline evaluations using random patches} on CIFAR-10.  Perturbations have $\ell_{\infty}$-norm bounded above by $16/255$, and poison budget is $1\%$ of training images.  Each number denotes an average (and standard error) over $24$ independent crafting and training runs along with randomly sampled source/target class pairs. Each run has a unique patch generated randomly.}
  \label{tab:bernoulli}
  \small
  \centering
  \begin{tabular}{lc}
    \textbf{Architecture}     
    & \textbf{ResNet-18}  
    \\
    \midrule
    Clean model val(\%)  & $92.16 \ (\pm 0.08)$\\
    Poisoned model val\ (\%) & $92.00\ (\pm 0.07)$\\
    Clean model source val\ (\%)      & $92.55 \ (\pm 0.98)$\\
    Poisoned model source val\ (\%)     & $91.77\ (\pm 1.09)$\\
    Poisoned model patched source val\ (\%)  & $\textbf{14.86}\ (\pm 5.06)$\\
    Attack Success  Rate\ (\%)       & $\textbf{82.05}\ (\pm 5.80)$\\
  \end{tabular}
\end{table*}

\subsection{Patch Size}
\label{app:patch_size}

To investigate the effect of patch size on the attack success rate, we perform the baseline evaluation discussed in Section \ref{baseline} using different patch sizes. From Table \ref{tab:patch_size}, we observe that by poisoning only $0.05\%$ of the training set and using a larger patch, we can effectively poison ImageNet. Furthermore, by using a proper amount of perturbation, Sleeper Agent works well with the smaller patches on both CIFAR-10 and ImageNet datasets. Visualizations of patched sources using different patch sizes are shown in Figure \ref{fig:imagenet_patch}.

\begin{table*}[h]
  \caption{\textbf{The effect of patch size}. Experiments are conducted on CIFAR-10 and ImageNet datasets with ResNet-18 models. Visualizations of different patched sources from ImageNet dataset can be found in Figure \ref{fig:imagenet_patch}. }
  
  \label{tab:patch_size}
  \centering
  \begin{adjustbox}{width=0.95\columnwidth,center}
  \begin{tabular}{cccccc}
    \textbf{Attack} & \textbf{Dataset}  & \textbf{Poison budget}  & \textbf{Patch size} &   \textbf{$\ell_{\infty}$-norm} &\textbf{Attack Success Rate} ($\%$)   \\
    \midrule
    Sleeper Agent\ (S = 1, T = 4) & CIFAR-10 & $1\%$ & $6\times6$   & $20/255$    &  $64.78$ \\
    Sleeper Agent\ (S = 1, T = 4) & CIFAR-10 & $1\%$ & $8\times8$   & $16/255$    &  $85.27$ \\
    \hline
    Sleeper Agent\ (S = 1, T = 2) & ImageNet & $0.05 \%$ & $25\times25$   & $16/255$    &  $38.00$ \\
    Sleeper Agent\ (S = 1, T = 2) & ImageNet & $0.05 \%$ & $25\times25$   & $24/255$    &  $52.00$ \\
    Sleeper Agent\ (S = 1, T = 2) & ImageNet & $0.05 \%$ & $30\times30$   & $16/255$    &  $44.00$ \\
    Sleeper Agent\ (S = 1, T = 2) & ImageNet   &  $0.05 \%$  & $45\times45$   & $16/255$ & $50.50$ \\

  \end{tabular}
  \end{adjustbox}
\end{table*}

\subsection{More Evaluations on ImageNet}

In addition to the experiments in Section \ref{baseline} and Appendix \ref{app:patch_size}, we provide more evaluations on ImageNet dataset focusing on low poison budget and smaller $\ell_{\infty}$-norm constraint. The evaluation results are listed in Table \ref{tab:imagenet_hard}. The results indicate that our proposed threat model is still effective by poisoning only 250 images in the ImageNet trainset. Additionally, under the hard $\ell_{\infty}$-norm constraint of $8/255$, Sleeper Agent has a partial success of one out of four (significantly better than random guess with a success rate of $0.001$ on ImageNet).

\begin{table*}[h]
  \caption{\textbf{ImageNet evaluations}. Experiments are conducted on ResNet-18 models.}
  
  \label{tab:imagenet_hard}
  \centering
  \begin{adjustbox}{width=0.95\columnwidth,center}
  \begin{tabular}{cccc}
    \textbf{Attack} & \textbf{Perturbation $\ell_{\infty}$-norm} & \textbf{Poison budget} & $\textbf{Attack Success Rate\ (\%)\ }$ \\
    \midrule
    Sleeper Agent\ (S = 1, T = 2) & $\textbf{8/255}$ & $0.05 \%$ (500 images) & $28.00 $ \\
    Sleeper Agent\ (S = 1, T = 2) & $16/255$ & \textbf{\textbf{$0.025 \%$} (250 images)} & $27.33 $ \\
  \end{tabular}
  \end{adjustbox}
\end{table*}

\subsection{Retraining Factor}

Table \ref{tab:retraining_factor} demonstrates the effect of the retraining factor on the attack success rate on the CIFAR-10 dataset. For $T$ larger than $4$, we do not see a considerable improvement in the attack success rate. Since increasing $T$ is costly, we choose $T=4$ as it simultaneously gives us a high success rate and is also significantly faster than $T=8$. We observe that even with $T=4$, the attack success rate is above $95\%$ in most trials

\begin{table*}[h]
  \caption{\textbf{Ablation studies on retraining factor}. Investigation of the effects of retraining factor $T$. Experiments are conducted on CIFAR-10 with ResNet-18 models, perturbations have $\ell_{\infty}$-norm bounded above by $16/255$, and the poison budget is $1\%$ of training images.}
  
  \label{tab:retraining_factor}
  \centering
  \begin{adjustbox}{width=0.5\columnwidth,center}
  \begin{tabular}{cc}
    \textbf{Retraining factor} & $\textbf{Attack Success Rate\ (\%)\ }$ \\
    \midrule
    $T = 0$ & $63.49 \ (\pm 6.13)$ \\
    $T = 2$ & $70.66 \ (\pm 6.66)$ \\
    $T = 4$ & $85.27 \ (\pm 5.90)$ \\
    $T = 8$ & $86.48 \ (\pm 6.26)$
  \end{tabular}
  \end{adjustbox}
\end{table*}

\subsection{Additional Defenses}
\label{app:add_def}

In addition to the defenses evaluated in Section \ref{subsec:defenses}, we evaluate Sleeper Agent on two recent defenses, ABL \citep{li2021anti} and ANP \citep{wu2021adversarial}. We evaluate against ANP with various threshold values, and we report the accuracy of a ResNet-18 on CIFAR-10, where Sleeper Agent poisons have $\ell_{\infty}$-norm bounded above by $16/255$.  We find that ANP cannot achieve a low attack success rate without greatly reducing the model’s validation accuracy on clean data.  Similarly, we find that ABL must decrease accuracy substantially to remove the backdoor vulnerability.  Even after 20 epochs of unlearning, Sleeper Agent has nearly a 60\% success rate. See Tables \ref{tab:rebuttal_anp} and \ref{tab:rebuttal_abl} for the complete results of these experiments.

\begin{table*}[h]
  \caption{\textbf{Adversarial Neuron Pruning (ANP)}.  We test Sleeper Agent against the defense, ANP \citep{wu2021adversarial}, across various threshold levels.  We see here that high threshold values can decrease the attack success rate but at a high cost in validation accuracy. Experiments are conducted on CIFAR-10 with ResNet-18 models.  Perturbations have $\ell_{\infty}$-norm bounded above by $16/255$.  
  }
  \label{tab:rebuttal_anp}
  \centering
  \small
  \begin{adjustbox}{width=0.7\columnwidth,center}
  \begin{tabular}{cccc}
    \textbf{Defense}  & \textbf{Threshold}   & \textbf{Validation Accuracy} ($\%$) & \textbf{Attack Success Rate} ($\%$)   \\
    \midrule
    None   &  -   &  $92.31$ & $85.27$ \\
    ANP   &  0.05   &  $80.05$ & $51.03$ \\
    ANP   &  0.10   &  $71.75$ & $27.87$ \\
    ANP   &  0.15   &  $50.47$ & $17.77$ \\
    ANP   &  0.20   &  $16.56$ & $3.35$ \\

  \end{tabular}
  \end{adjustbox}
\end{table*}

\begin{table*}[h]
  \caption{\textbf{Anti-Backdoor Learning (ABL)}.  We test Sleeper Agent against the defense, ABL \citep{li2021anti}, across various numbers of unlearning epochs.  We see here that many unlearning epochs can decrease the attack success rate but at a high cost in validation accuracy. Experiments are conducted on CIFAR-10 with ResNet-18 models.  Perturbations have $\ell_{\infty}$-norm bounded above by $16/255$.  
  }
  \label{tab:rebuttal_abl}
  \centering
  \small
  \begin{adjustbox}{width=0.8\columnwidth,center}
  \begin{tabular}{cccc}
    \textbf{Defense}  & \textbf{Unlearning Epochs}   & \textbf{Validation Accuracy} ($\%$) & \textbf{Attack Success Rate} ($\%$)   \\
    \midrule
    None   &  -   &  $92.31$ & $85.27$ \\
    ABL   &  5   &  $87.53$ & $70.72$ \\
    ABL   &  10   &  $86.85$ & $68.35$ \\
    ABL   &  15   &  $82.34$ & $64.11$ \\
    ABL   &  20   &  $64.55$ & $59.51$ \\

  \end{tabular}
  \end{adjustbox}
\end{table*}

\begin{table*}[ht!]
  \caption{\textbf{Adversarial Training}.  We test Sleeper Agent against adversarial training via PGD with a perturbation radius of $4/255$.  Experiments are conducted on CIFAR-10 with ResNet-18 models.  Poison perturbations have $\ell_{\infty}$-norm bounded above by $16/255$.  
  }
  \label{tab:rebuttal_adversarial_training}
  \centering
  \small
  \begin{adjustbox}{width=0.65\columnwidth,center}
  \begin{tabular}{ccc}
    \textbf{Defense}  & \textbf{Validation Accuracy} ($\%$) & \textbf{Attack Success Rate} ($\%$)   \\
    \midrule
    None   &  $92.31$ & $85.27$ \\
    Adv. Training    &  $88.63$ & $52.83$ \\

  \end{tabular}
  \end{adjustbox}
\end{table*}

\section{Implementation Details}
\label{app:imp_det}

\subsection{Experimental Setup}

\begin{figure}[ht!]
    \centering
    \includegraphics[width=0.10\linewidth]{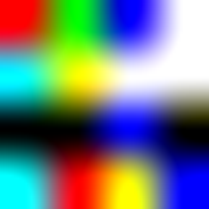}
    \caption{The trigger we use in our primary experiments.}
    \label{fig:trigger}
\end{figure}

The most challenging setting for evaluating a backdoor attack targets victim models that are trained from scratch \citep{schwarzschild2020just}. On the other hand, it is crucial to compute the average attack success rate on all patched source images in the validation set to evaluate effectiveness reliably. Hence, to evaluate our backdoor attack, we first poison a training set using a surrogate model as described in Algorithm \ref{alg1}, then the victim model is trained in a standard fashion on the poisoned training set from scratch with random initialization. After the victim model is trained, to compute the \emph{attack success rate}, we measure the average rate at which patched source images are successfully classified as the target class. To be consistent and to provide a fair comparison to \citet{saha2019hidden}, in our primary experiments, we use a random patch selected from \citet{saha2019hidden} as shown in Figure \ref{fig:trigger}. In our baseline experiments, following \citet{saha2019hidden}, the patch size is $8\times8$ for CIFAR-10 ($6.25\%$ of the pixels) and $30\times30$ for the ImageNet ($1.79\%$ of the pixels). Note that the choice of the patch in our implementation is not essential, and our model is effective across randomly selected patches (see Appendix \ref{app:patch_choice}). More experiments on smaller patch sizes are presented in Appendix \ref{app:patch_size}.

\subsection{Models and Hyperparameters}

For our evaluations, we use ResNet-18, ResNet-34, MobileNet-v2, and VGG11 \citep{he2016deep, sandler2018mobilenetv2, simonyan2014very}. For training ResNet-18 and ResNet-34, we use initial learning rate $0.1$, and for MobileNet-v2 and VGG11, we use initial learning rate $0.01$. We schedule learning rate drops at epochs $14$, $24$, and $35$ by a factor of $0.1$. For all models, we employ SGD with Nesterov momentum, and we set the momentum coefficient to $0.9$. We use batches of 128 images and weight decay with a coefficient of $4 \times 10^{-4}$. For all CIFAR-10 experiments, we train and retrain for $40$ epochs, and for validation, we train the re-initialized model for 80 epochs. For the ImageNet experiments, we employ pre-trained models from \verb+torchvision+ to start crafting, and for retraining and validation, we apply a similar procedure explained: training for 80 epochs for both retraining and validation while we schedule learning rate drops at epochs 30, 50, and 70 by a factor of 0.1. To incorporate data augmentation, for CIFAR-10, we apply horizontal flips with probability $0.5$ and random crops of size $32 \times 32$ with zero-padding of $4$. And for the ImageNet, we use the following data augmentations: 1) resize to $256 \times 256$, 2) central crop of size $224 \times 224$, 3) horizontal flip with probability $0.5$, 4) random crops of size $224 \times 224$ with zero-padding of $28$. 

\subsection{Implementation of Benchmark Experiments}

In Section \ref{sec:comp_others} we compared our threat model with Clean-Label Backdoor \citet{turner2019label} and Hidden-Trigger Backdoor \citet{saha2019hidden}. For both methods, We follow the same procedure used in their papers as described in \citet{schwarzschild2020just}. Specifically, to reproduce the clean-label attack, we use the implementation code provided in \citet{schwarzschild2020just}. To get each poison, we compute the PGD-based adversarial perturbation to each image, and then the trigger is added to the image \citep{schwarzschild2020just, turner2019label}.

\subsection{Defense Details}
In Section \ref{subsec:defenses}, we test a variety of defenses against our proposed attack. For filtering-based defenses, such as SS, AC, we craft the poisons as usual (according to Algorithm \ref{alg1}). We then train a model on the poisoned data. After this, we apply one of the selected defenses to identify what training data may have been poisoned. We then remove the detected samples, and retrain a *second* network from scratch on the remaining data. Finally, we evaluate the attack success rate (on the backdoored class) using this second network. For STRIP, we simply apply the defense at test time for the first network, and filter out any patched images that exceed an entropy threshold in their predictions. In this case, an attack is considered a success if a backdoored input is not detected at test time, *and* misclassified as the target class.

\subsection{Runtime Cost}

\begin{figure*}[t!]
    \centering
    \includegraphics[width=0.85\linewidth]{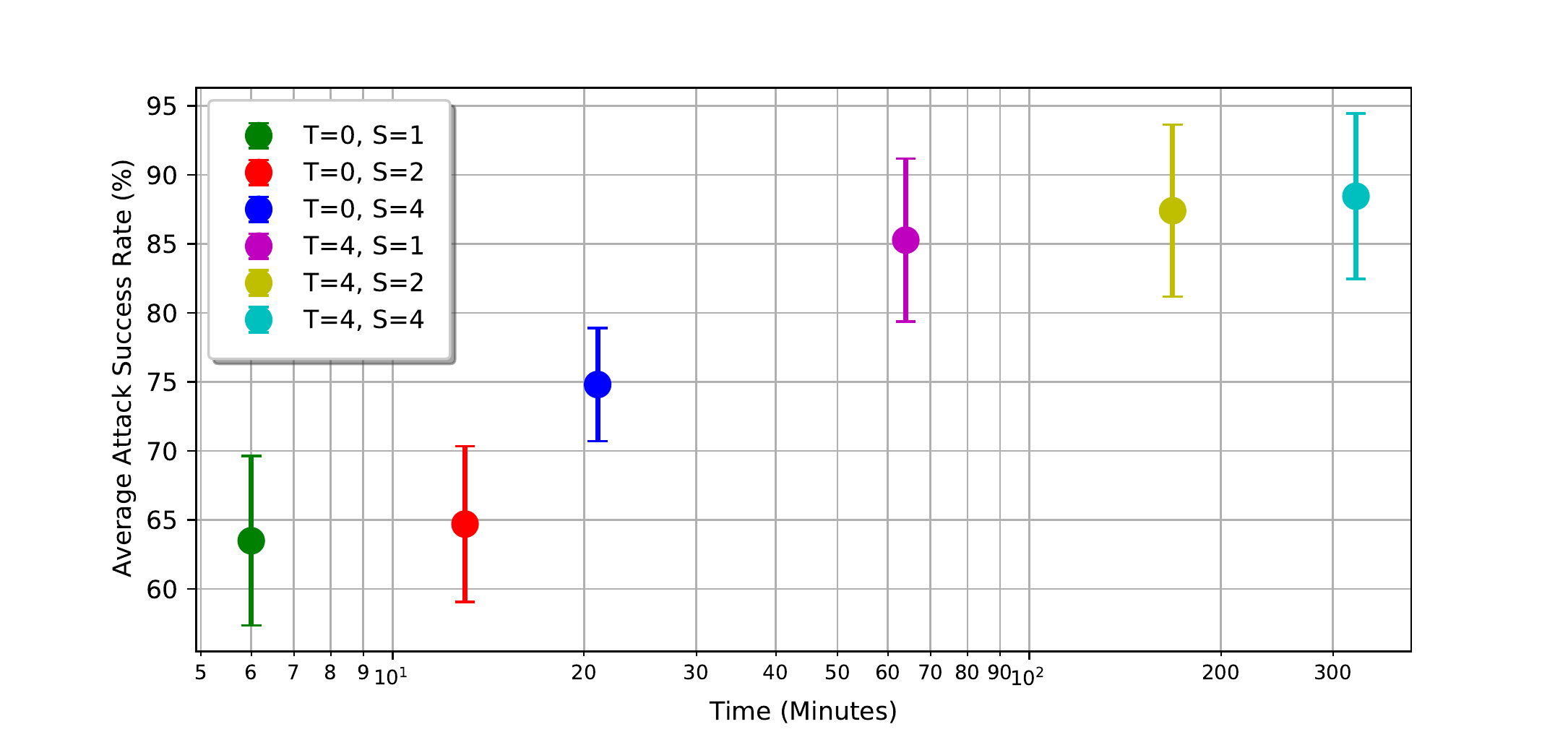}
    \caption{Average poisoning time for various Sleeper Agent setups. All experiments are conducted on CIFAR-10 with ResNet-18 models. Perturbations have $\ell_{\infty}$-norm bounded above by $16/255$, and the poison budget is $1\%$ of training images. $T$ denotes the training factor and $S$ denotes the ensemble size.}
    \label{fig:time}
\end{figure*}

\begin{figure*}[t!]
    \centering
    \includegraphics[width=0.85\linewidth]{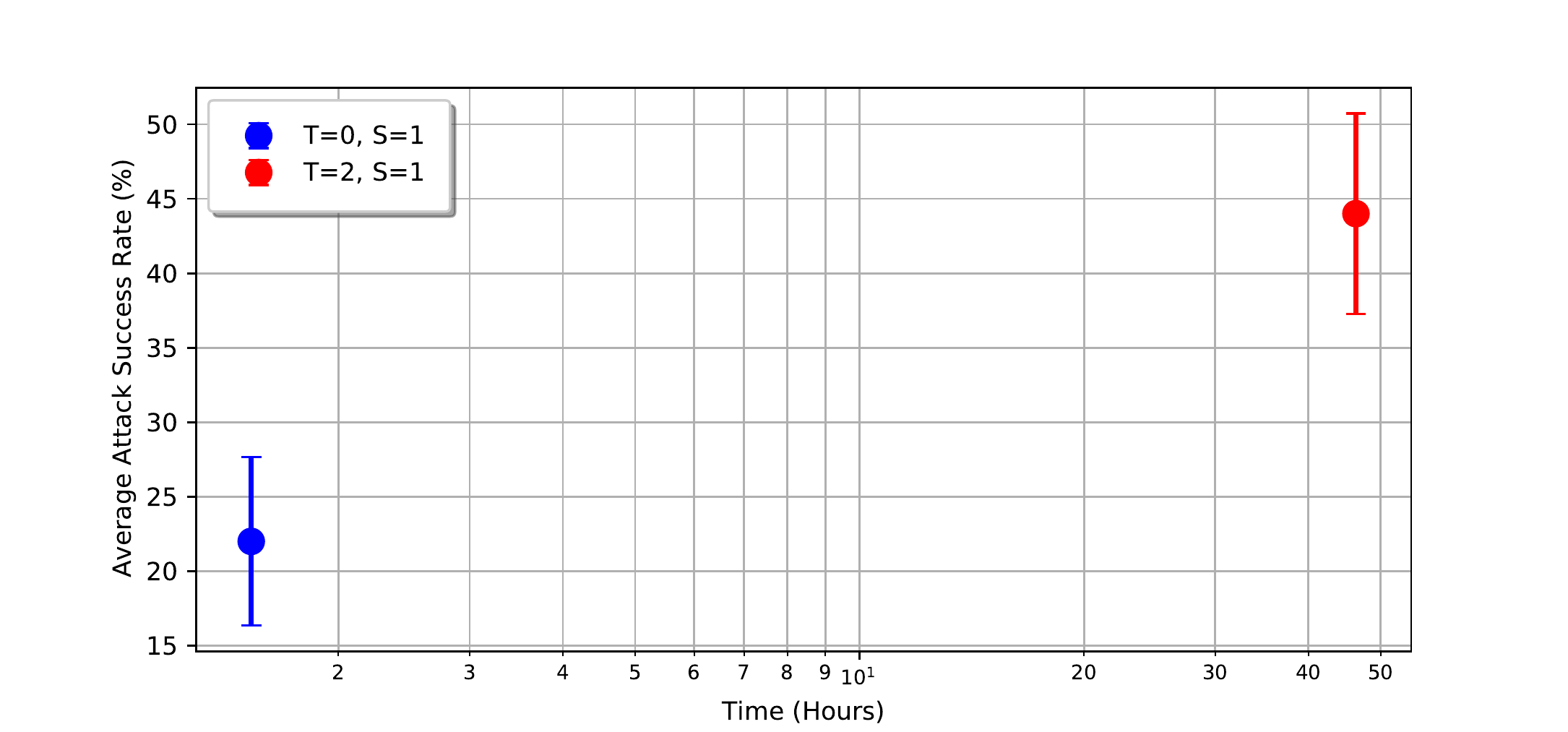}
    \caption{Average poisoning time for various Sleeper Agent setups. All experiments are conducted on ImageNet with ResNet-18 models. Perturbations have $\ell_{\infty}$-norm bounded above by $16/255$, and the poison budget is $0.05\%$ of training images. $T$ denotes the training factor and $S$ denotes the ensemble size.}
    \label{fig:time_imagenet}
\end{figure*}

We use two NVIDIA GEFORCE RTX 2080 Ti GPUs for baseline evaluations on CIFAR-10 and two to four NVIDIA GEFORCE RTX 3090 GPUs for ImageNet baseline evaluations depending on the network size. Figures \ref{fig:time} and \ref{fig:time_imagenet} show the time cost of Sleeper Agent with different settings.

\section{Visualizations}
\label{app:vis}

\begin{figure}[h]
    \centering
    \captionsetup[subfigure]{justification=centering, belowskip=0pt}
    \begin{subfigure}[t]{\columnwidth}
        \raisebox{-\height}{\includegraphics[width=0.99\textwidth]{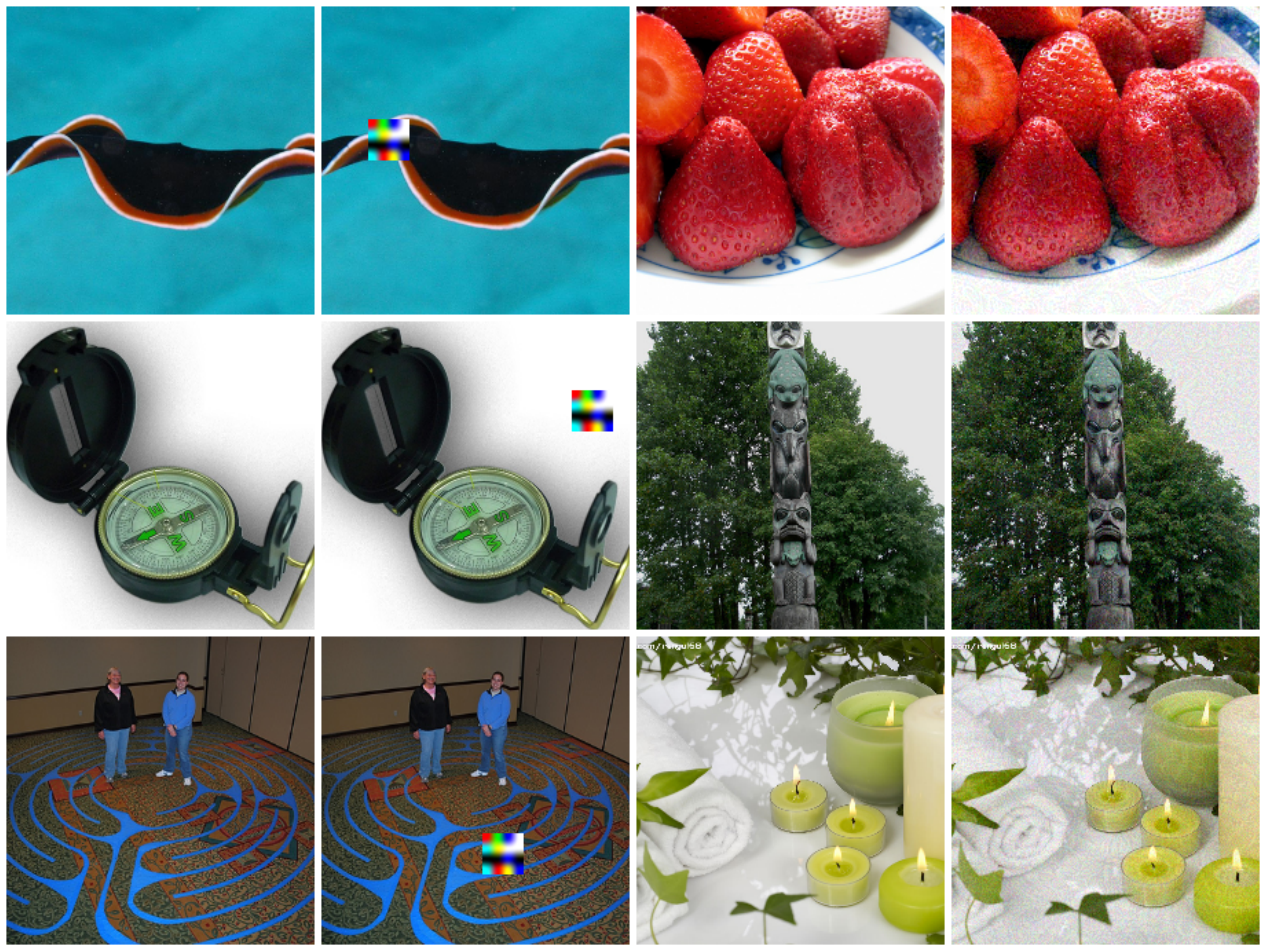}}
    \end{subfigure}
    \begin{subfigure}[t]{\columnwidth}
        \raisebox{-\height}{\includegraphics[width=0.99\textwidth]{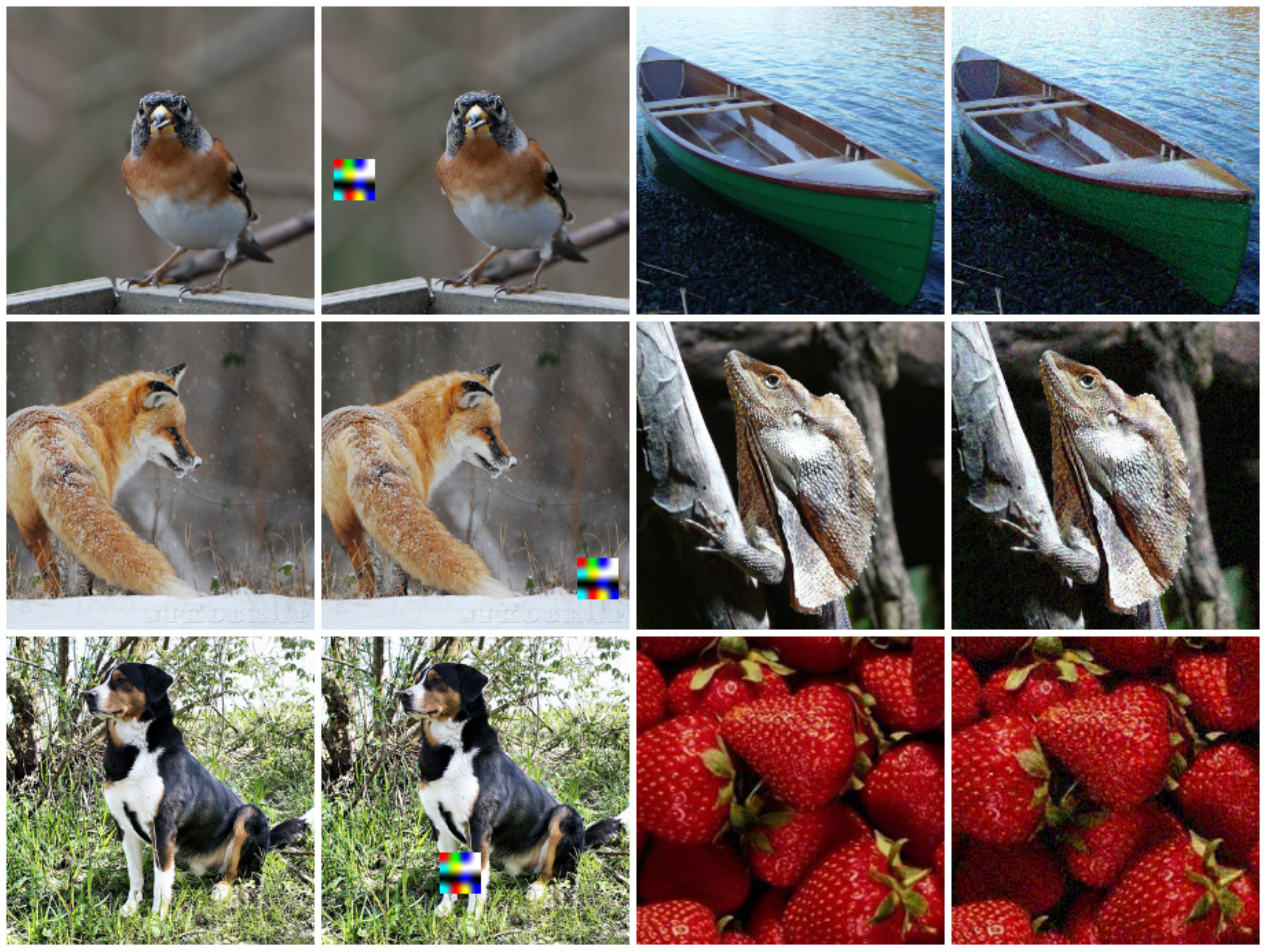}}
    \end{subfigure}

    \caption{Visualizations of the successful attacks on the ImageNet dataset. Each row includes the clean source, patched source, clean target, and poisoned target, respectively. Perturbations have $\ell_{\infty}$-norm bounded above by $16/255$, and the patch size is $30$.} 
    \label{fig:imagenet_vis}
    
\end{figure}

\begin{figure}[ht]
    \centering
    \includegraphics[width=0.75\linewidth]{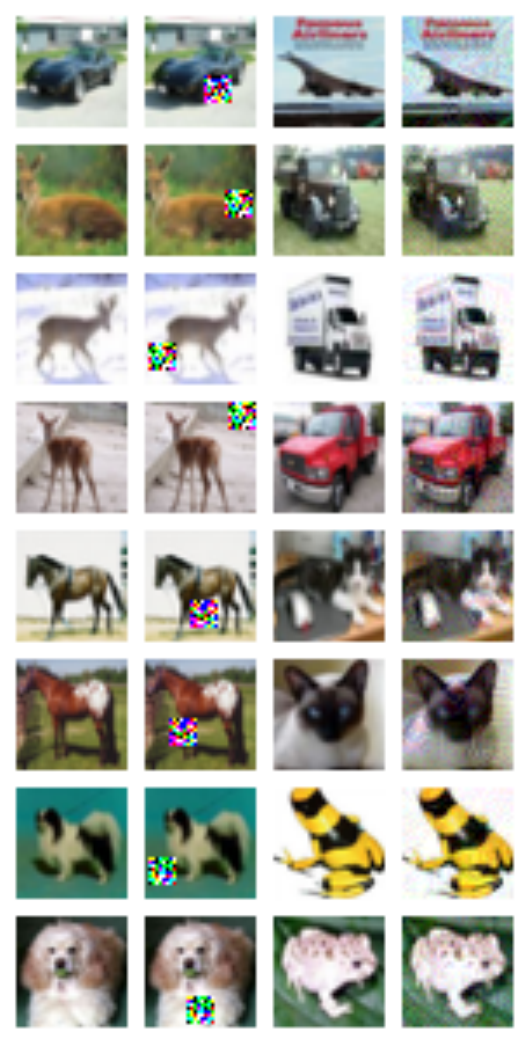}
    \caption{Visualizations of the successful attacks on the CIFAR-10 dataset. Each row includes the clean source, patched source, clean target, and poisoned target,  respectively. Perturbations have $\ell_{\infty}$-norm bounded above by $16/255$ and the patch size is $8$. Here, patches are randomly generated as described in Appendix \ref{app:patch_choice}.}
    \label{fig:cifar_vis_random_patch}
\end{figure}

\begin{figure*}[ht]
    \centering
    \captionsetup[subfigure]{justification=centering, belowskip=0pt}
    \begin{subfigure}[t]{\columnwidth}
        \centering
        \raisebox{-\height}{\includegraphics[width=0.9\textwidth]{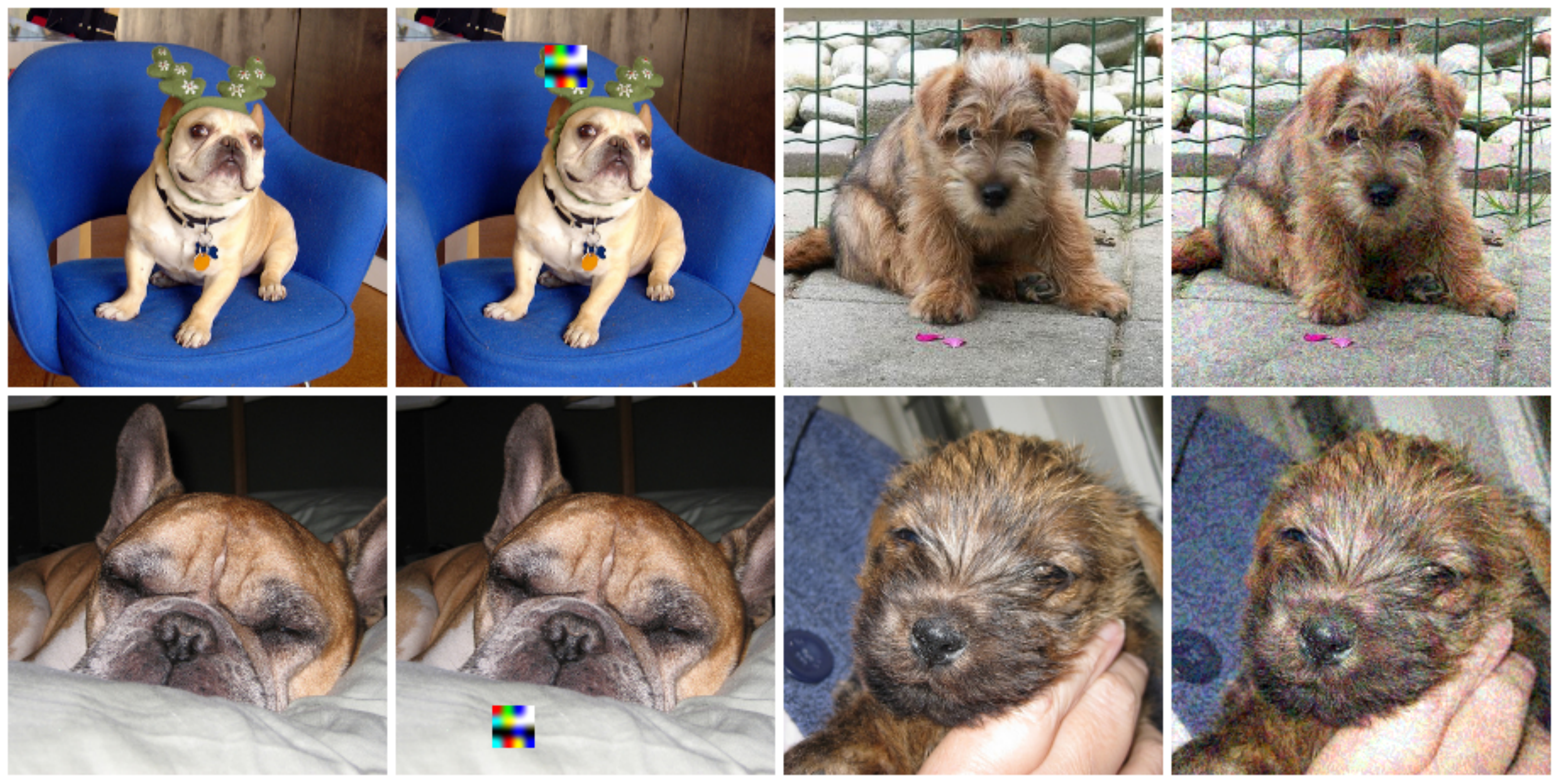}}
        \caption{Patch size $= 25$}
    \end{subfigure}
    \begin{subfigure}[t]{\columnwidth}
        \centering
        \raisebox{-\height}{\includegraphics[width=0.9\textwidth]{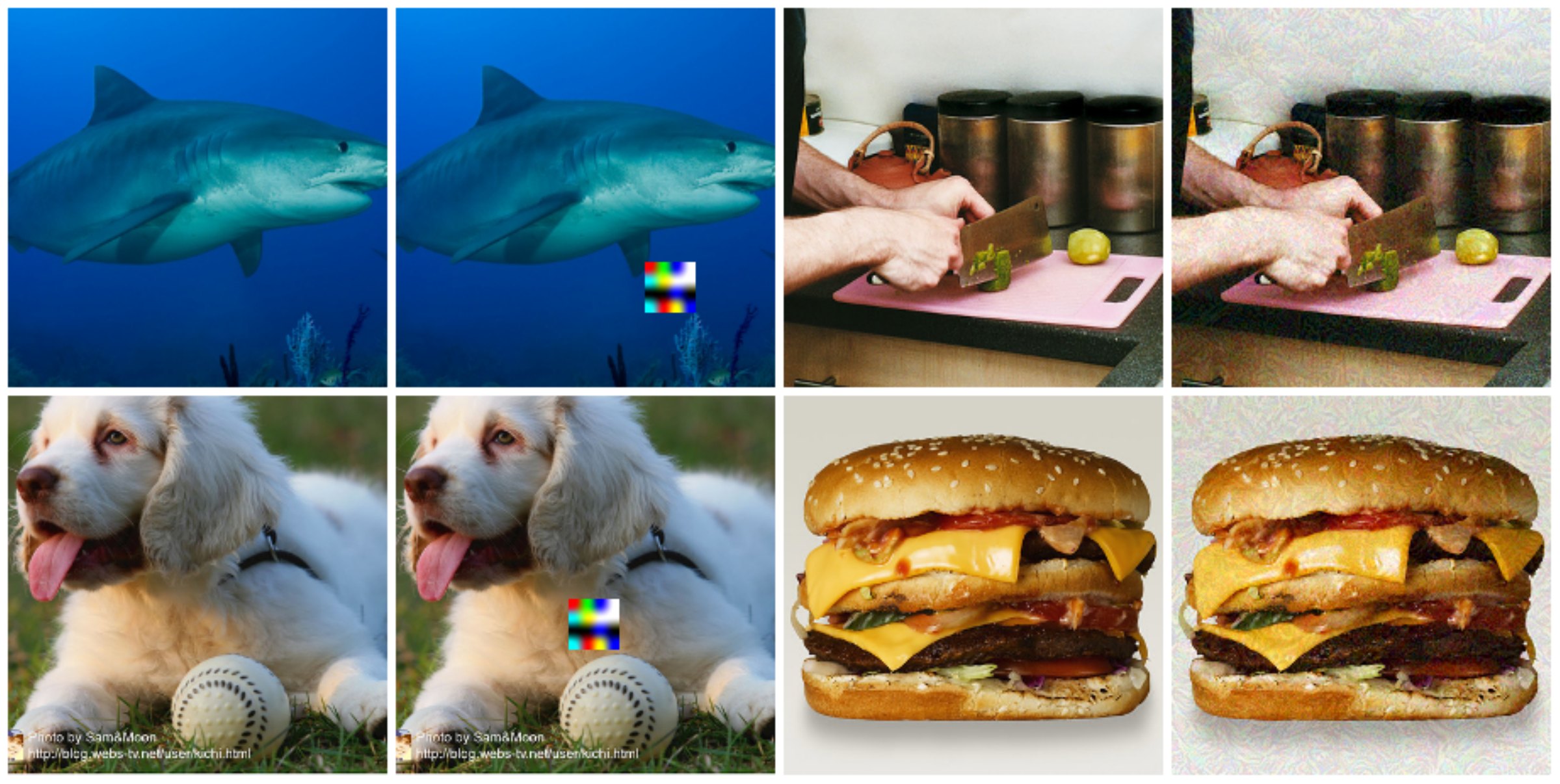}}
        \caption{Patch size $= 30$ (baseline)}
    \end{subfigure}
    \begin{subfigure}[t]{\columnwidth}
        \centering
        \raisebox{-\height}{\includegraphics[width=0.9\textwidth]{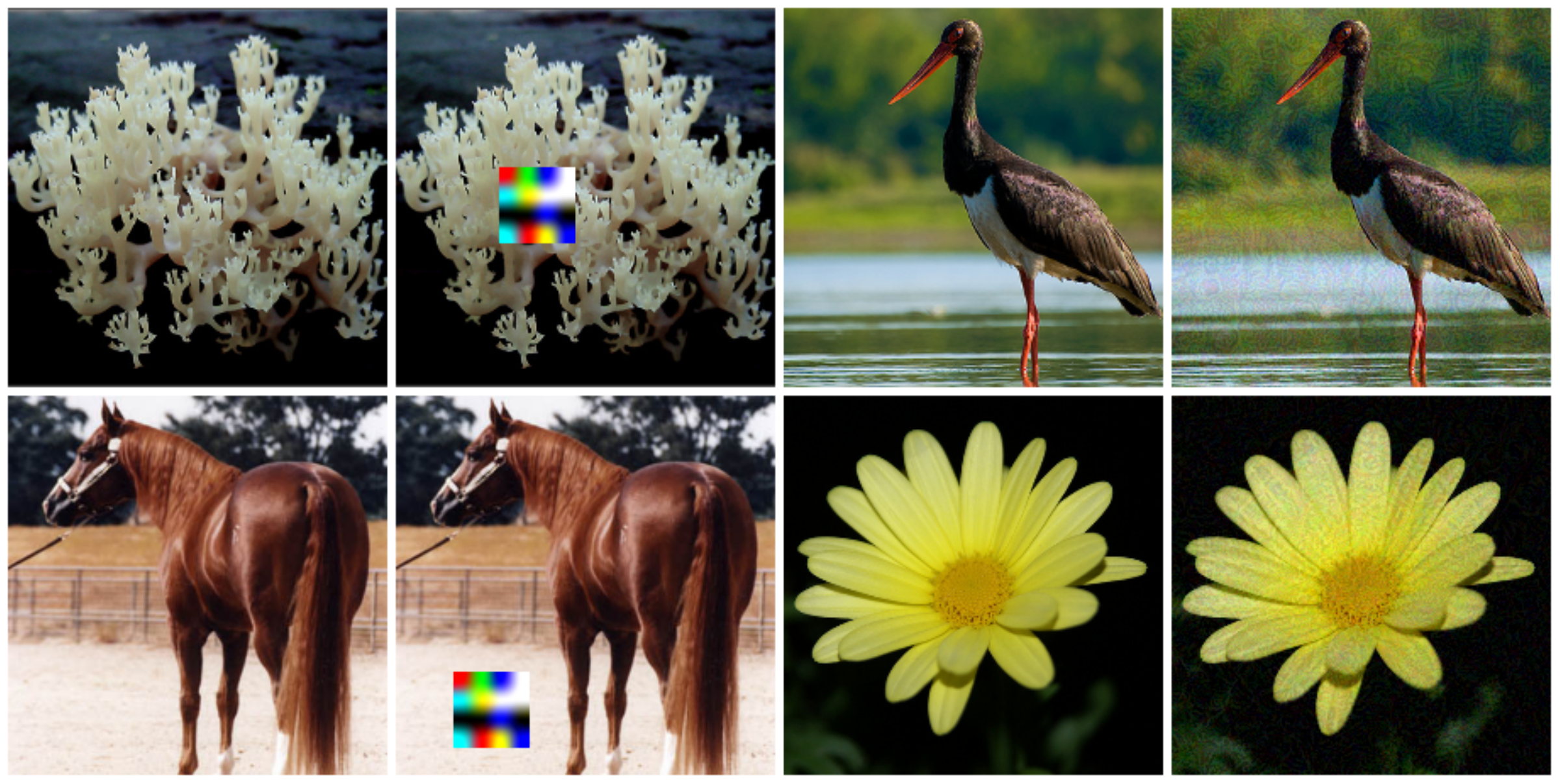}}
        \caption{Patch size $= 45$}
    \end{subfigure}
    
    \caption{Sample clean source (first column), patched source (second column), clean target (third column), and poisoned target (fourth column) from the ImageNet dataset with different trigger size. Perturbations have $\ell_{\infty}$-norm bounded above by $16/255$.} 
    \label{fig:imagenet_patch}
    
\end{figure*}

\begin{figure*}[t!]
    \centering
    \captionsetup[subfigure]{justification=centering, belowskip=0pt}
    \begin{subfigure}[t]{\columnwidth}
        \centering
        \raisebox{-\height}{\includegraphics[width=0.5\textwidth]{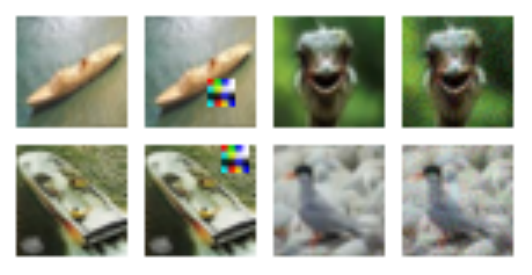}}
        \caption{$\epsilon = 8$}
    \end{subfigure}
    \begin{subfigure}[t]{\columnwidth}
        \centering
        \raisebox{-\height}{\includegraphics[width=0.5\textwidth]{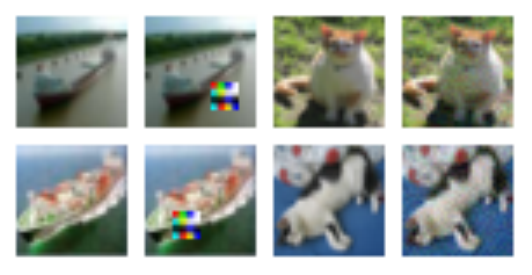}}
        \caption{$\epsilon = 10$}
    \end{subfigure}
    \begin{subfigure}[t]{\columnwidth}
        \centering
        \raisebox{-\height}{\includegraphics[width=0.5\textwidth]{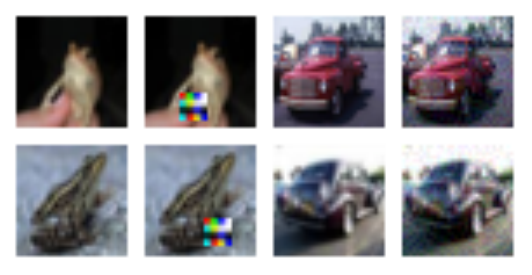}}
        \caption{$\epsilon = 12$}
    \end{subfigure}
    \begin{subfigure}[t]{\columnwidth}
        \centering
        \raisebox{-\height}{\includegraphics[width=0.5\textwidth]{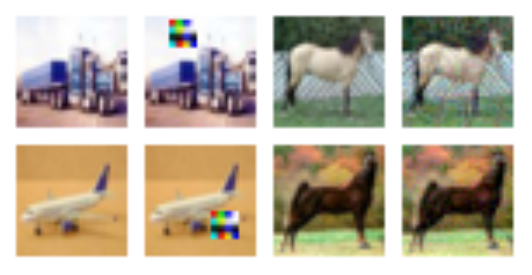}}
        \caption{$\epsilon = 14$}
    \end{subfigure}
    \begin{subfigure}[t]{\columnwidth}
        \centering
        \raisebox{-\height}{\includegraphics[width=0.5\textwidth]{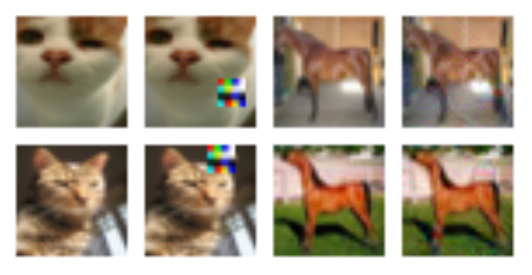}}
        \caption{$\epsilon = 16$}
    \end{subfigure}
    
    \caption{Sample clean source (first column), patched source (second column), clean target (third column), and poisoned target (fourth column) from the CIFAR-10 dataset with different $\ell_{\infty}$-norm perturbation.} 
    \label{fig:cifar10_epsilon}
    
\end{figure*}

\begin{figure*}[t!]
    \centering
    \captionsetup[subfigure]{justification=centering, belowskip=0pt}

    \begin{subfigure}[t]{0.2\columnwidth}
        \raisebox{-\height}{\includegraphics[width=0.99\textwidth]{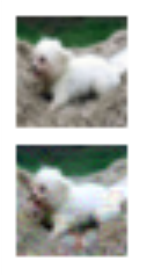}}
        \caption{$\epsilon = 16$}
    \end{subfigure}
    \begin{subfigure}[t]{0.2\columnwidth}
        \raisebox{-\height}{\includegraphics[width=0.99\textwidth]{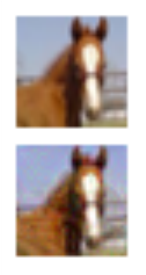}}
        \caption{$\epsilon = 14$}
    \end{subfigure}
    \begin{subfigure}[t]{0.2\columnwidth}
        \raisebox{-\height}{\includegraphics[width=0.99\textwidth]{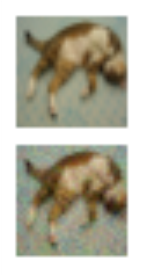}}
        \caption{$\epsilon = 10$}
    \end{subfigure}
    \begin{subfigure}[t]{0.2\columnwidth}
        \raisebox{-\height}{\includegraphics[width=0.99\textwidth]{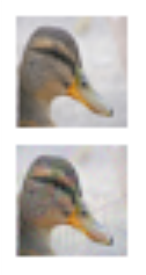}}
        \caption{$\epsilon = 8$}
    \end{subfigure}
    
    \caption{Visualization of clean targets (first  row) and poisoned targets (second row) with different $\ell_{\infty}$-norms from the CIFAR-10 dataset.} 
    \label{fig:cifar_epsilon_target}
    
\end{figure*}

In this section, we present more visualizations of the successful attacks on CIFAR-10 and ImagNet datasets. Figures \ref{fig:imagenet_vis}, \ref{fig:cifar_vis_random_patch},  \ref{fig:imagenet_patch}, \ref{fig:cifar10_epsilon}, and \ref{fig:cifar_epsilon_target} show patched sources and poisoned targets generated by Sleeper Agent on CIFAR-10 and ImageNet. We observe that the generated perturbed images and their corresponding clean images are hardly distinguishable by the human eye, especially in the last column of Figure \ref{fig:cifar_epsilon_target} where the $\ell_{\infty}$-norm of perturbation is bounded above by $8/255$.

\end{document}